\pdfoutput=1

\documentclass[11pt]{article}

\usepackage{acl}

\usepackage{times}
\usepackage{latexsym}
\usepackage{amssymb}
\usepackage{amsmath}
\usepackage{multicol}
\usepackage{multirow}
\usepackage{enumitem}

\usepackage{amssymb}
\usepackage{pifont}

\newcommand{\jo}[1]{\textcolor{brown}{#1}}
\newcommand{\LCG}[1]{\textsc{LCGScore}}

\usepackage{graphicx}
\usepackage{booktabs}
\usepackage{comment}
\usepackage{tikz}
\usetikzlibrary{shapes.geometric}
\usepackage{lingmacros}

\usepackage[T1]{fontenc}

\usepackage{csquotes}

\usepackage{microtype}

\usepackage{fancyvrb}
\usepackage{verbatimbox}
\DefineVerbatimEnvironment{VerbOutB}{Verbatim}{fontfamily=arial}

\usepackage{pifont}
\definecolor{cadmiumgreen}{rgb}{0.0, 0.42, 0.24}
\newcommand{\cmark}{\textcolor{cadmiumgreen}{\ding{51}}}%
\newcommand{\xmark}{\textcolor{red}{\ding{55}}}

\definecolor{palechestnut}{rgb}{1.0, 0.3, 0.3}
\definecolor{lavender(web)}{rgb}{0.9, 0.9, 0.98}

\usepackage{chngcntr}
\title{A Dynamic,  Interpreted 
\textit{CheckList}
for Meaning-oriented 
NLG Metric Evaluation -- through the Lens of Semantic Similarity Rating
}

\author{Laura Zeidler \and Juri Opitz \and Anette Frank \\
         Department of Computational Linguistics\\ Heidelberg University, Germany\\
         \texttt{\{zeidler|opitz|frank\}@cl.uni-heidelberg.de} }

\begin{document}
\maketitle

\begin{abstract}
Evaluating the quality of
generated text is difficult, since traditional NLG evaluation metrics, focusing more on surface form than meaning, often fail to assign appropriate scores.
This is especially problematic for AMR-to-text evaluation, given the abstract nature of AMR.
Our work aims to support the development and improvement of NLG evaluation metrics that focus on \textit{meaning},
by developing a  \textit{dynamic CheckList} for NLG metrics that is \textit{interpreted} by being organized around meaning-relevant linguistic phenomena.
Each test instance consists of a pair of sentences
with their AMR graphs and
a human-produced 
\textit{textual semantic similarity}  or \textit{relatedness} score. 
Our \textit{CheckList} facilitates comparative evaluation of metrics 
and 
reveals strengths and weaknesses of novel and traditional metrics.
We demonstrate the usefulness of \textit{CheckList} by designing a new 
metric \textsc{GraCo} that computes
lexical cohesion graphs over AMR concepts.
Our analysis suggests that \textsc{GraCo} 
presents an interesting NLG
metric worth future investigation and that 
meaning-oriented NLG metrics can 
profit from graph-based metric components using AMR.
\end{abstract}

\section{Introduction}

Abstract Meaning Representation (AMR, \citet{banarescu-etal-2013-abstract}) has become popular in NLP, one of the reasons being that AMR captures the essence of a sentence's meaning, while abstracting away from syntactic idiosyncrasies. Especially AMR-to-text generation  \citep{konstas-etal-2017-neural,song-etal-2018-graph,wang-etal-2020-amr,blloshmi-etal-2021-spring} has received much attention 
for applications that require text generation from structured content. However, the evaluation of text generated from AMR has been argued to be unsatisfactory \citep{manning-etal-2020-human}. Also,
\citet{opitz-frank-2021-towards} show that the syntactic diversity of sentences generated from AMR 
is challenging for traditional NLG metrics, 
especially when candidates differ from the reference in surface properties.

Several metrics have been proposed that aim to rate the similarity of the meaning of sentences or phrases
(\citet{BERTscore}; \citet{opitz-frank-2021-towards}; \citet{zhao-etal-2019-moverscore}).
However,
it is difficult to judge where exactly such a metric fails, making it hard for developers to further improve it. 
To address similar problems, \citet{ribeiro-etal-2020-beyond} recently proposed a "task-agnostic methodology for testing NLP models" called \textit{CheckList}.
They argue that such a method should be used for testing NLP systems instead of
solely relying on
automatic metrics, which can overestimate a model’s performance. Similar processes have been applied in early NLP research, e.g. with the TSNLP testsuite \citep{lehmann-etal-1996-tsnlp}.
Inspired by \textit{CheckList}, in this work we aim to build a testsuite to enable systematic study and development of NLG evaluation metrics, with a focus on meaning.

\begin{figure}[t!]
\centering
\resizebox{\linewidth}{!}{
    \centering
    \includegraphics{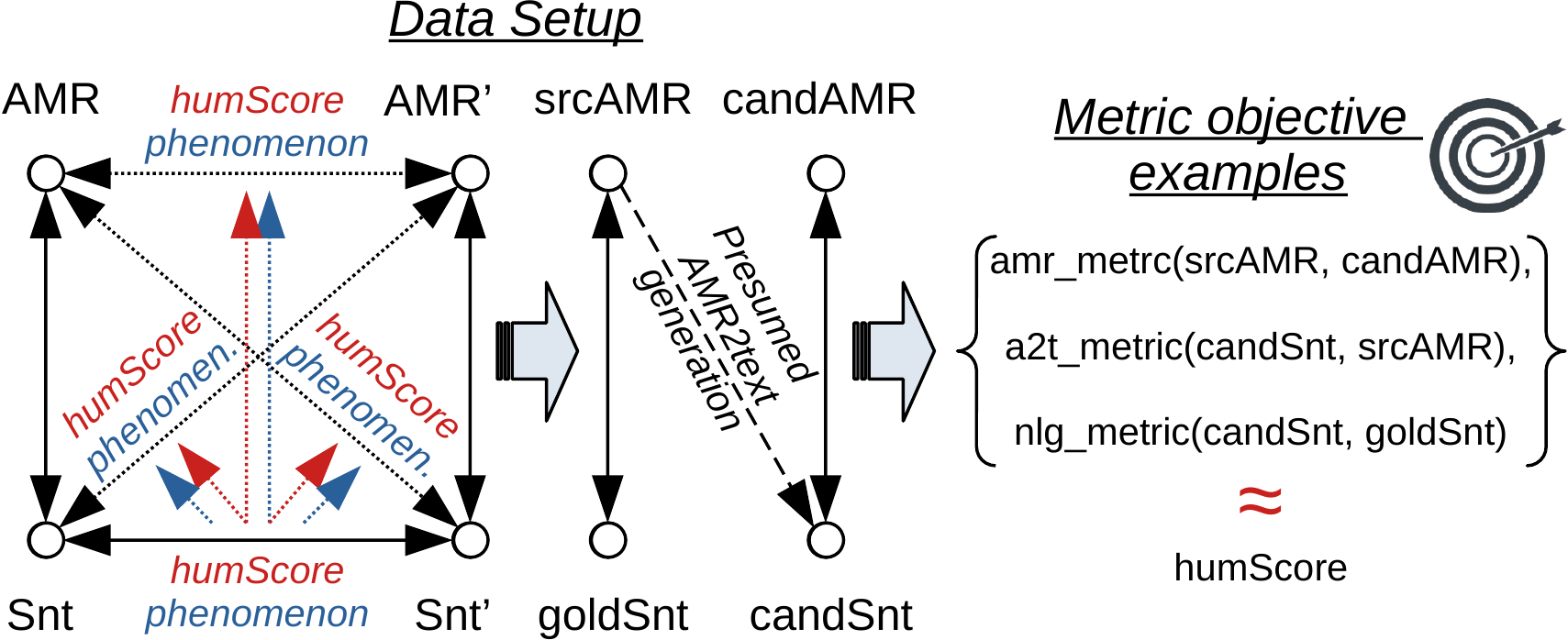} 
    }
    \caption{
    Our \textit{CheckList} design for evaluating 
    meaning-oriented NLG metrics against human semantic textual similarity and relatedness judgements -- applicable to textual, meaning graph based and hybrid metrics.
    }
    \label{fig:eval}
\end{figure}

Given the high variability of surface realizations that can be mapped into a single AMR graph, building reliable AMR-to-text NLG evaluation metrics is hard. Hence, it can be useful to construct a systematic \textit{CheckList}, organized around diverse linguistic properties, to measure the performance of different metrics in an interpretable way.
We frame our proposed \textsc{CheckList}\footnote{
The term \textit{CheckList}, coined by \citet{ribeiro-etal-2020-beyond}, refers to their proposed methodology as well as concrete instantiations of such testsuites. We thus use the term \textit{CheckList} (in italics), to refer to our interpreted NLG testsuite.
} and analyses derived from it in an AMR-to-Text NLG setting, and focus especially on a metric's capability to assess how well a specific meaning component of an AMR is reflected in its  textual realization.  We measure this using sentence pairs that differ in single linguistic aspects
and measure how well various 
NLG metrics are able to rate such  
meaning differences. We compare the metric scores to human judgments from semantic textual similarity (STS) and relatedness datasets and analyze the metrics using our interpreted \textit{CheckList} (an outline is shown in Fig.\ \ref{fig:eval}).
Our contributions in this work are as follows: 

\begin{enumerate}[label=\roman*), noitemsep]
    \item We empirically identify properties  relevant for rating the quality of generated sentences based on their meaning.
    \item We design an extensible, interpreted \textit{CheckList} for evaluating NLG metrics, which offers 939 paired sentences with human judgements, covering 11 core linguistic phenomena.
    \item We propose a new 
    metric \textsc{GraCo} to assess the semantic similarity of sentence pairs through the lens of AMR graphs. 
    \item To showcase the potential of our approach, we provide an extensive comparative analysis of different types of NLG metrics,  measuring their capacity of rating sentence similarity and relatedness according to linguistic differences.
\end{enumerate}

\section{Related Work}

\paragraph{AMR-to-text evaluation} 
Systems generating text from AMR graphs are typically evaluated using NLG metrics that were
originally designed for other NLG tasks. \textsc{Bleu} \cite{papineni-etal-2002-bleu} or the \textsc{chrF(++)}  \cite{stanojevic-etal-2015-results,popovic-2015-chrf, popovic-2016-chrf, popov-2017-word} metrics, e.g., are extensively used in MT.
But \citet{may-priyadarshi-2017-semeval} have shown that \textsc{Bleu} does not correspond well to human ratings of generations from AMR. Confirming this result,
\citet{manning-etal-2020-human} argue 
that existing automatic metrics fail to provide nuanced views on AMR-to-text generation quality. 
In an attempt to
mitigate such issues,
\citet{opitz-frank-2021-towards}
introduced a metric that combines meaning ($\cal{M}$) and form ($\cal{F}$) assessment  in a  
weighted $\cal{MF}$ score, finding that
system performances
differ considerably in these two key quality aspects.

 But to date,
 little is known about
 how 
 different metrics measure meaning differences of generated sentences with regard to specific meaning alterations that may 
 occur between a source and a reference. Our work provides a method and resources that can be used for performing such a
 detailed 
 assessment for AMR-to-text generation metrics, and NLG evaluation metrics in general.

\paragraph{Checklist} 
The current practice for evaluating NLP models is to 
assess their performance on  
unseen test data.
Yet,
summarizing performance in a single numerical score makes it difficult to assess where a model fails and 
how to fix remaining errors
\citep{wu-etal-2019-errudite}. 
\citet{ribeiro-etal-2020-beyond} therefore proposed \textsc{CheckList}, a 
methodology and
tool for evaluating NLP systems based on the idea of \textit{behavioural testing}, 
often used in 
software engineering. 
It aims at assessing
specific capabilities of a system
by 
testing whether 
inputs that feature specific properties will produce the expected output, without requiring knowledge of system's inner workings.
This procedure is well-known in NLP,  
where before the rise of large-scale evaluation datasets, systems were tested and evaluated on so-called \textit{testsuites} \citep{lehmann-etal-1996-tsnlp} that focused on
specific \textit{linguistic capabilities}.
\citet{ribeiro-etal-2020-beyond} adopted this approach
to make their methodology applicable to many different NLP tasks. 
They  evaluate multiple models on 
Sentiment Analysis, QA or Machine Reading Comprehension,
showing that their
method 
is
beneficial in NLP:
complementary to broad-scale evaluations, it can reveal specific points of failure, 
hence giving more detailed insight into a model's performance.

\paragraph{Semantic Textual Similarity (STS)}
Judging the 
similarity of texts is essential 
in tasks such as IR, text summarization or QA.
But capturing semantic ambiguity, syntactic variance and paraphrasing is difficult.
Hence, research started to investigate
\textit{Semantic} Textual Similarity (STS)\footnote{STS is a main component of SentEval and follow-up challenges, initiated by \citet{conneau-kiela-2018-senteval}.}, by tasking systems to judge the semantic similarity of sentences.
Besides knowledge-based 
and distributional methods,
 neural methods have recently been proposed for STS estimation \citep{10.1145/3440755}.
For example, S(entence)-BERT \citep{reimers-gurevych-2019-sentence} leverages
pre-trained language models 
to predict STS scores,
building on the insight of models that compute general sentence representations using paired sentence encoders \cite{conneau-etal-2017-supervised}.
These models outperform most traditional STS metrics, but 
lack interpretability.
In our work we leverage 
STS and SentEval challenge datasets with human-rated semantic similarity (STS) and semantic relatedness (SICK) scores, to construct an interpreted \textit{CheckList} that can be used to assess meaning-oriented NLG evaluation metrics, by evaluating them against human ratings. 

\section{An Interpreted 
Testsuite for Meaning- oriented NLG Evaluation Metrics}

\subsection{Aims and Method}
\label{sec:aims}

The challenge of AMR-to-text NLG evaluation lies in the wide variability of sentences that can 
verbalize an  abstract meaning representation. In our \textit{CheckList}, 
we will consider human judgements of semantic textual similarity as a criterion for evaluating the adequacy of different NLG metrics for the AMR-to-text NLG evaluation task. 

\begin{figure}
    \centering
    \resizebox{0.8\linewidth}{!}{
\begin{tikzpicture}
\scriptsize
    \node[shape=rectangle,draw=black,thick] (D) at (0,1.5) {scrAMR};
    \node[shape=ellipse,draw=blue,thick,align=center] (A) at (1,1) {xv0 /\\ hit-01};
    \node[shape=ellipse,draw=blue,thick,align=center] (B) at (0,-0.5) {xv2 /\\ boy};
    \node[shape=ellipse,draw=blue,thick,align=center] (C) at (2,-0.5) {xv1 /\\ baseball}; ;
\scriptsize
    \path [->] (A) edge node[left] {$ARG0$} (B);
    \path [->](A) edge node[right] {$ARG1$} (C);
    
    \node[draw] at (0,-1.5) {goldSnt};
    \node[text width=2.7cm] at (0.9,-2) {A boy is hitting a baseball};
\end{tikzpicture}

\hspace{1mm}
\begin{tikzpicture}
\scriptsize
\node[shape=rectangle,draw=black,thick] (D) at (0,1.5) {candAMR};
    \node[shape=ellipse,draw=green,thick,align=center] (A) at (1,1) {xv0 /\\ hit-01};
    \node[shape=ellipse,draw=green,thick,align=center] (B) at (0,-0.5) {xv2 /\\ child};
    \node[shape=ellipse,draw=green,thick,align=center] (C) at (2,-0.5) {xv1 /\\ baseball}; ;

    \path [->] (A) edge node[left] {$ARG0$} (B);
    \path [->](A) edge node[right] {$ARG1$} (C);
    
    \node[draw] at (0,-1.5) {candSnt};
    \node[text width=2.8cm] at (0.9,-2) {A child is hitting a baseball};
\end{tikzpicture}
}
    \caption{Example of a test case in our \textit{CheckList} consisting of two sentence and AMR pairs. Drawn from the SICK dataset, with semantic relatedness score 4.4.}
    \label{fig:example}
\end{figure}
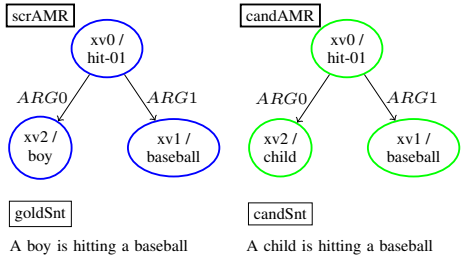

Specifically, we employ sentence pairs with human scores from the SICK and STS benchmarks\footnote{\scriptsize{\url{https://github.com/facebookresearch/SentEval}}} as test instances for our \textit{CheckList} (cf.\ Fig.\ \ref{fig:example}).
We select pairs that \textit{differ} by specific phenomena that can affect their semantic similarity, such as additional modifiers of a noun or verb, negation, or changes in the 
semantic roles of verb arguments.
We parse such sentence pairs $S_{A,B}$ into pairs of AMR graphs $AMR_{A,B}$ that we manually validate. 

Given such instances, we consider sentences $S_A$ and $S_B$ as a reference and candidate generation, and a pair of $AMR$ and $S$ as a sentence generated from an input AMR. For $AMR_A$ we can take $S_A$ as gold reference and $S_B$ as a candidate generation;  conversely, $S_B$ can serve as a reference for $AMR_B$, and $S_A$ as a candidate. We then interpret the human score for  $S_{A,B}$ as a gold standard for a metric score that rates the appropriateness of $S_B$ for $AMR_A$, given $S_A$ as a reference, or $S_A$ for $AMR_B$, given $S_B$ as reference
(see Fig.\ \ref{fig:eval}).

Following this rationale, our \textit{CheckList} will offer curated input AMR graphs, their underlying sentences as references, and paired sentences from STS or SICK data points as candidate generations.
The human 
scores serve as an objective to assess and compare various NLG evaluation metrics for their suitability in  (A)MR-to-text  evaluation tasks. 

\paragraph{Aims} 
Our \textit{CheckList} is intended as a tool for researchers to build new or assess existing NLG metrics, regarding their ability to assess specific meaning aspects by comparing them to human judgements,
thereby helping users to improve metrics, or better understand differences between metrics in meaning-oriented NLG evaluation in general and  AMR-to-text generation in particular.

The suite is \textit{interpreted} in two ways: by structuring the instances according to linguistic phenomena, and by pairing each sentence with its AMR graph, so that sentences can be compared at the textual and at the meaning representation level.
Finally, the \textit{CheckList} is conceived to be  \textit{dynamic}, by inviting developers to add new linguistic phenomena, test cases, and metrics.

\paragraph{Method} To achieve this, we proceed as follows:

\textbf{i) Empirical investigation} We investigated sentences generated from 
the 'Little Prince Corpus'\footnote{\url{https://amr.isi.edu/download.html}} using the AMR-to-text system of \citet{song-etal-2018-graph}.
We studied differences between the original and the generated sentences, to determine core phenomena that may influence the semantic similarity judgement of sentences generated from AMR towards their references. We distilled a list of phenomena shown in Table \ref{phenomena} that we further extended with phenomena observed in the STS and SICK datasets.

\begin{table}[t!]
\resizebox{0.92\linewidth}{!}{
\small{
\begin{tabular}{@{}p{1.7cm}@{}|p{3.7cm}|p{4.0cm}@{}}
\toprule
Pheno- menon& Reference &AMR-to-text Generation\\
\midrule
Antonymy & Flowers are so inconsistent !& flowers are so consistent .\\[1ex]
Negation &My Drawing Number One .& not my picture number one .\\[1ex]
Omission &the prince laughed , puzzled . &the prince laughed .\\[1ex]
Passive &The wind blows them away . &they were blown away by wind .\\[1ex]
Role Switch& The planet was inhabited by a
conceited man .&
the conceit man is inhabited by
the planet .\\[1ex]
\midrule
\parbox[t]{1.7cm}{more\\phenomena}&\multicolumn{2}{p{6cm}}{hyponymy, co-hyponymy, partial synonymy, \linebreak articles,  subordinate
clause types}\\
\bottomrule
\end{tabular}}}
\caption{(Modified) sentence pairs from  
AMR-to-text 
on the Little Prince AMR corpus.}
\label{phenomena}
\end{table}

\textbf{ii) Selection from STS and SICK} Next, we select instances from the STS and Semantic Relatedness datasets (\S 5.1) that exhibit the phe\-no\-mena identified in \textbf{i)}, and establish a suite of sentence pairs  with their assigned human scores and respective AMRs. The data is structured into subsets exhibiting single phenomena, 
and is organized as an extensible \textit{CheckList}.

\textbf{iii) NLG metric scores \& evaluation} 
We implement scorers for various NLG metrics, and provide code to evaluate them via multiple measures
to 
assess their strengths and weaknesses 
in view of phenomena captured in the CheckList.
In addition, we propose a novel  metric \textsc{GraCo} (\S \ref{sec:metrics}) that 
constructs lexical cohesion graphs over tokens represented in the sentence's AMR, and compare it to existing metrics. The full range of functionalities to investigate NLG metrics is embedded into a \textsc{CheckList} design \cite{ribeiro-etal-2020-beyond} (cf.\ \ref{checklist_descr}).

\textbf{iv) Analysis and Interpretation} We analyze the results and show how our \textit{CheckList} enables systematic assessment of strenghts and weaknesses of NLG metrics when applied to outputs of AMR-to-text systems, taking into account the nature of  different metrics in view of different phenomena.

\subsection{Textual and AMR-based 
metrics}
\label{sec:metrics}

With our \textit{CheckList} we aim at the evaluation of diverse 
metrics used in NLG and in semantic parsing, which we structure along two dimensions (cf.\ Table \ref{tab:extendedMetric_categorization}):
metrics that evaluate candidate generations based on a) their textual (\textit{tM}) vs.\ graph (\textit{gM}) representations or both (hybrid, \textit{hyM}), and b) whether the metric is based on symbolic as opposed to embedding representations. 
We don't include trained metrics, since their interpretation is difficult and would go beyond the current scope, but they can
be evaluated on our \textit{CheckList}, too. Table \ref{tab:metrics} provides an overview of characterizing traits of these metric types, which we will refer to in our analyses in \S \ref{sec:evaluation}.

\paragraph{Word/Char Ngram Matching Metrics}

Originally developed for MT evaluation, 
the \textsc{Bleu} \citep{papineni-etal-2002-bleu}, Meteor \citep{lavie-agarwal-2007-meteor} and chrF++ \citep{popovic-2015-chrf}
metrics 
have been increasingly used for evaluating NLG systems by comparing generated text to a reference on textual symbols. \textsc{Bleu} and Meteor 
compute overlap in word ngrams, while chrF++ extends
the character ngram metric chrF by adding
word ngrams.

\if false
\begin{table*}[hbt!]
    \centering
    \scalebox{0.76}{
    \begin{tabular}{llrrr}
         & & \multicolumn{3}{c}{dependency on gold structure} \\
      metric category & metric & goldSnt & candAMR & srcAMR (always given)  \\
        graphMetric & \{S($^2$)match, WLK, WWLK\} & n & y & y \\
        graphMetric$^{parse}$ & see above w/ parsed candSnt & n & n & y \\
        graphMetric$^{parse}_{parse}$ & see above w/ parsed candSnt \& goldSnt & y & n & n \\
        textMetric & \{BERTscore, Meteor, BLEU, chrF++\}& y & n & n \\
        hybridMetric & \textsc{GraCo} (this paper) & y & y & y \\
    \end{tabular}}
    \caption{Metric categorization and dependencies on gold structure types.}
    \label{tab:extendedMetric_categorization}
\end{table*}
\fi

\begin{table}[t!]
\resizebox{\linewidth}{!}{
    \begin{tabular}{@{}l@{~}l@{}c@{~}c@{~}c@{}}
    \toprule
         & & \multicolumn{3}{@{}c@{}}{gold information} \\
         category & metric & gldS & cndAMR & srcAMR \\
         \midrule
        $gM$ & S($^2$)match, W(W)LK & n & y & y \\
        $gM^{cndS}$ & S($^2$)match, W(W)LK  & n & n & y \\
        $gM^{cndS}_{gldS}$ & S($^2$)match, W(W)LK  & y & n & n \\
        $tM$ & BERTsc, Meteor, BLEU, chrF++& y & n & n \\
        $hyM$ & \textsc{GraCo} (this paper) & y & y & y \\
        \bottomrule
    \end{tabular}}
    \caption{Categorization of metrics into graph-based \textit{gM}, text-based
    \textit{tM} and hybrid \textit{hyM} metrics, and their dependencies on gold information.
    }
    \label{tab:extendedMetric_categorization}
\end{table}

\paragraph{Embedding-based Metrics}

\textsc{BERTScore},
proposed by
\citet{BERTscore}, 
allows for reference-based evaluation using dense representations.
Reference and candidate sentences are embedded with \textsc{Bert} to obtain  contextualized representations for each token. A mapping 
between candidate and reference tokens is computed by greedy
matching, based on
cosine similarity of the encoding vectors.
\textsc{BERTS}core shows a high correlation with human judgements for 
MT and Image Captioning tasks \citep{BERTscore}. But while the metric is 
clearly meaning-based, it is focused on lexical meaning, and is not well equipped to capture word order 
and compositional meaning.

\paragraph{AMR Parse Evaluation Metrics}
While the previous metrics evaluate candidates against a reference at the \textit{textual level} ($tM$), in our CheckList, 
we complement them by assessing  similarity of meaning \textit{structurally}, at the level of AMR graphs constructed from candidate and reference ($gM$).

We distinguish
three potential setups: i) the metric is computed on manually rectified gold graphs ($gM$ in Table \ref{tab:extendedMetric_categorization}); ii) an 
integrated parser component 
constructs an automatic candidate AMR \textit{cndAMR} from the candidate sentence \textit{cndSnt} to alleviate the requirement for a golden \textit{cndAMR} ($gM^{cndS}$ in Table \ref{tab:extendedMetric_categorization}); iii.) the 
parser 
constructs both 
\textit{srcAMR} and \textit{candAMR} from the reference and candidate sentence, i.e., we trade the dependency on a golden \textit{srcAMR} against the dependency on a golden reference sentence ($gM^{cndS}_{gldS}$ in Table \ref{tab:extendedMetric_categorization}). Variants ii) and iii) have also been used in the $\mathcal{M}$ (`Meaning') component of MF-score \cite{opitz-frank-2021-towards}. For simplicity, in this paper, we assume access to gold graphs and only consider $gM$, $tM$, and $hyM$ metrics.

As AMR graph metrics, we use the canonical \textsc{Smatch}
\citep{cai-knight-2013-smatch}, the recent \textsc{S$^2$Match} metric proposed by \citet{opitz-etal-2020-amr}, and Weisfeiler-Leman based AMR graph similarity proposed by \citet{opitz-etal-2021-weisfeiler} that match contextualized AMR graphs. 

\textsc{Smatch} is a \textit{binary} triple overlap metric that assesses the structural similarity of candidate and reference AMRs, where a triple is a pair of AMR nodes connected by a labeled edge. \textsc{S$^2$match}, by contrast, 
computes a \textit{graded} triple overlap score using the embedding similarity between the concept nodes of a triple pair, to reflect concept similarity in the overall AMR similarity score. Given a reference AMR for \textit{'a kitten meows'}, \textsc{S$^2$match} will assign a relatively high score for a candidate AMR for \textit{'a cat meows'} that  reflects high lexical similarity of \textit{kitten} and \textit{cat} in the overall 
score, while \textsc{Smatch} will assign it a much lower score.

The Weisfeiler-Leman 
AMR metric comes in two variants: W(eisfeiler)L(eman)K(ernel) (\textsc{Wlk})
compares contextualized AMR graphs structurally, 
while
W(asserstein)WLK (\textsc{WWlk})
compares the contextualized AMR graphs in 
latent space, using
an alignment-based Wasserstein distance. \textsc{WWlk} extends \textsc{S$^{2}$Match} beyond the lexical level, to capture 
\textit{compositional} meaning similarity at the phrasal level, as between \textit{'a young cat meows'} vs.\ \textit{'a kitten meows'}.

\paragraph{Hybrid Metrics} The above metrics 
take as input sentence pairs or AMR pairs. But a mea\-ning-\-orien\-ted NLG metric may profit from con\-si\-der\-ing both explicit meaning structure as captured in AMR, and the textual level, to leverage knowledge from pretrained language models trained on text.
We thus propose a \textbf{hybrid similarity metric} \textsc{GraCo}, which
is based on 
\textit{Lexical Cohesion Graphs} proposed by \citet{Sporleder-li-2009-unsupervised}. They construct an undirected graph 
from a 
text sequence where each node represents a content word, and 
compute edge weights  
between the lexical nodes 
using
Normalized Google Distance \cite{cilibrasi2007google}. 
By averaging the weights they derive a
\textit{connectivity} score for the graph. 
In their work they use
the lexical cohesion graph of
a given token sequence to predict whether it has an \textit{idio\-matic} as opposed to a \textit{literal} meaning, depending on whether the presence of its subgraph in the overall graph raises or lowers the overall connectivity score.

\begin{figure}[t!]
\centering
\resizebox{0.7\linewidth}{!}{
    \centering
    \includegraphics{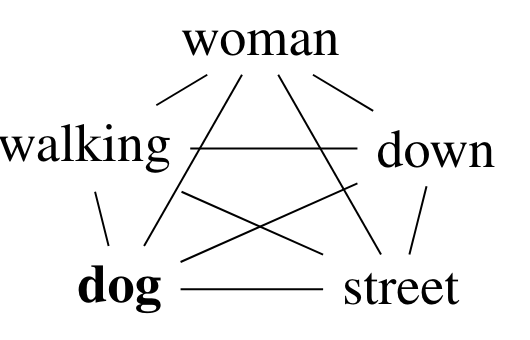} \includegraphics{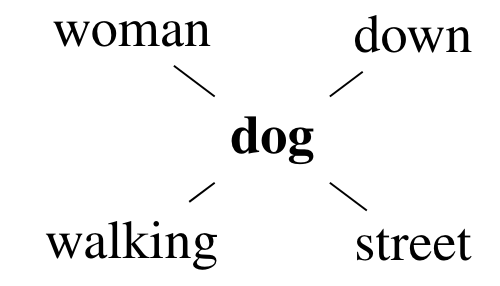}
    }
\resizebox{0.8\linewidth}{!}{
     \includegraphics{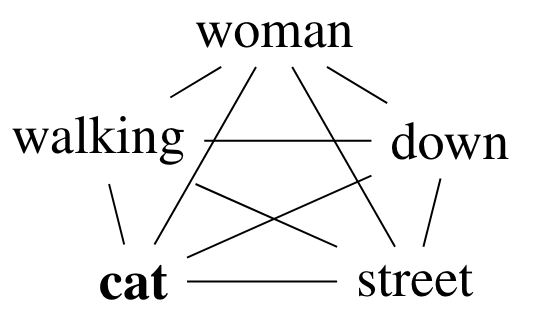} \includegraphics{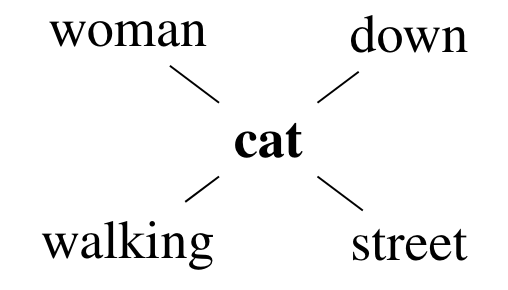}
     }
\if false
\begin{tikzpicture}
    \node[] (D) at (0.5,2.75) {woman};
    \node[] (A) at (-0.75,2) {walking};
    \node[] (B) at (-0.5,1) {\textbf{dog}};
    \node[] (C) at (1.75,2) {down};
    \node[] (E) at (1.5,1) {street}; ;

    \path [-] (A) edge node[left] {} (B);
    \path [-] (A) edge node[right] {} (C);
    \path [-] (A) edge node[left] {} (D);
    \path [-] (A) edge node[left] {} (E);
    \path [-] (B) edge node[right] {} (C);
    \path [-] (B) edge node[right] {} (D);
    \path [-] (B) edge node[right] {} (E);
    \path [-] (C) edge node[right] {} (D);
    \path [-] (C) edge node[right] {} (E);
    \path [-] (D) edge node[right] {} (E);
\end{tikzpicture}
\hfill
\begin{tikzpicture}
    \node[] (D) at (0.5,2.5) {woman};
    \node[] (A) at (0.5,1) {walking};
    \node[] (B) at (1.5,1.75) {\textbf{dog}};
    \node[] (C) at (2.5,2.5) {down};
    \node[] (E) at (2.5,1) {street}; ;

    \path [-] (A) edge node[left] {} (B);
    \path [-] (B) edge node[right] {} (C);
    \path [-] (B) edge node[right] {} (D);
    \path [-] (B) edge node[right] {} (E);
\end{tikzpicture}
\hfill\\\vspace{3mm}
\begin{tikzpicture}
    \node[] (D) at (0.5,2.75) {woman};
    \node[] (A) at (-0.75,2) {walking};
    \node[] (B) at (-0.5,1) {\textbf{cat}};
    \node[] (C) at (1.75,2) {down};
    \node[] (E) at (1.5,1) {street}; ;

    \path [-] (A) edge node[left] {} (B);
    \path [-] (A) edge node[right] {} (C);
    \path [-] (A) edge node[left] {} (D);
    \path [-] (A) edge node[left] {} (E);
    \path [-] (B) edge node[right] {} (C);
    \path [-] (B) edge node[right] {} (D);
    \path [-] (B) edge node[right] {} (E);
    \path [-] (C) edge node[right] {} (D);
    \path [-] (C) edge node[right] {} (E);
    \path [-] (D) edge node[right] {} (E);
\end{tikzpicture}
\hfill
\begin{tikzpicture}
    \node[] (D) at (0.5,2.5) {woman};
    \node[] (A) at (0.5,1) {walking};
    \node[] (B) at (1.5,1.75) {\textbf{cat}};
    \node[] (C) at (2.5,2.5) {down};
    \node[] (E) at (2.5,1) {street}; ;

    \path [-] (A) edge node[left] {} (B);
    \path [-] (B) edge node[right] {} (C);
    \path [-] (B) edge node[right] {} (D);
    \path [-] (B) edge node[right] {} (E);
\end{tikzpicture}
\fi
    \caption{Two lexical cohesion graphs: fully connected (left) and reduced
    (right) for sentences $S_A$: \textit{The woman is walking the \textbf{dog} down the street} -- $S_B$: \textit{The woman is walking the \textbf{cat} down the street}.}
    \label{fig:star}
\end{figure}
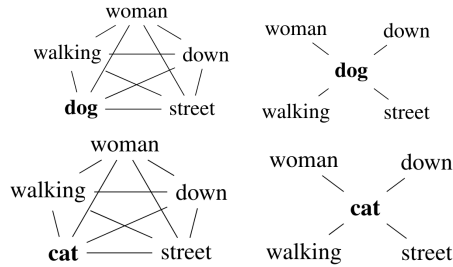

We adapt \newcite{Sporleder-li-2009-unsupervised}'s approach to define a \textit{hybrid metric} that measures the similarity of sentence pairs via their AMR graphs. 
We do this by building a lexical cohesion graph from the concept nodes present in a sentence's AMR. To do so, we align words from the sentence with concepts in the AMR graph using the JAMR \cite{flanigan-etal-2014-discriminative} alignment tool. The concepts are either represented using contextualized \textsc{BERT} embeddings
or pretrained GloVe 
word 
embeddings.
To compute edge weights, we follow \citet{haagsma-etal-2018-side} and compute cosine similarity between nodes. We pursue two strategies.
i) We follow \citet{Sporleder-li-2009-unsupervised} and compute cosine similarity between all possible pairs of nodes of a single graph, creating a \textit{fully connected}
graph. 
Alternatively, ii) we compute a reduced graph that only takes into account edges 
connecting nodes that \textit{differ} between the two sentences and their respective graphs 
(see Fig.\ \ref{fig:star}).
In case graph $g_A$ differs from graph $g_B$ in a single
concept which is only present in $g_A$, the reduced graph $g_B$ is empty, and we assign a connectivity score of $1$ (consistent with anything).

By applying this method to a pair of sentences 
$S_A$ and $S_B$, we obtain their \textit{connectivity scores} $cs_A$ and $cs_B$, 
the average of their respective graphs' edge weights.
From these we compute the \textsc{GraCo} Score (\ref{eq:lcg}) that rates the similarity of 
$S_A$ and $S_B$ by taking  
the difference between $cs_A$ and $cs_B$ to model their semantic difference  
-- which we convert to a similarity score by  subtracting it  
from 1. \vspace{-1mm}
\begin{equation}\label{eq:lcg}
    \textsc{GraCo}Score = 1 - | cs_A - cs_B |
\end{equation}
\noindent
The resulting metric is hybrid by relying on the sentence's \textit{AMR}  to select text tokens for 
the
connectivity graph -- and represents nodes with \textit{contextualized embeddings}  
in the \textsc{BERT}
variant. 

\section{Semantic Phenomena}

We consider  structural and lexical phenomena
that are likely to affect a sentence's meaning.
Details and example AMRs are given in Appendix \ref{A:phenomena}.\footnote{AMR specifications follow 
\citet{amr_guidelines}.} 

\subsection{Structural Phenomena} \label{sub_phen}
\textbf{Aspect} Given its abstract nature, AMR does not represent aspect, hence present perfect and simple present are not distinguished in an AMR graph\footnote{This phenomenon was only found in the STS data.}.\\
\textbf{Negation}
AMR represents negation with the feature \texttt{:polarity -}.
Fig.\ \ref{fig:neg1} (\ref{A:negation}) shows sen\-tence negation, with
\texttt{polarity} attached to 
the matrix verb. 
Fig.\ \ref{fig:neg2} (\ref{A:negation})
shows an AMR that ne\-gates a 
constituent in a 
sentence. Both 
verb- and 
constituent negation
are represented in the testsuite.

\textbf{Omission or Hallucination} 
of words or phrases is a recurring
problem in NLG 
\citep{xiao-wang-2021-hallucination} especially 
for AMR-to-text 
\citep{manning-etal-2020-human}. 
We sampled three types involving
\textit{adjectives}, \textit{adverbs}, \textit{PPs}.
In AMR, omission/hallucination is captured by (non-)existence of the corresponding structure (see Fig.\ \ref{fig:om1}, \ref{A:hallucination}).

\textbf{Passive}
AMR does not distinguish active from passive voice: AMR graphs for active vs.\ passive sentences do not differ and do not reflect voice. 

\textbf{Semantic 
Role Switch}
describes cases
where two verb arguments 
switch 
semantic 
roles.
Fig.\ \ref{fig:ssrs} (\ref{A:roleswitch})
shows that
the switch
changes the \texttt{:ARG} roles of both arguments, involving two triples.

\textbf{Subordinate Clauses}
In AMR, relative clauses can involve \textit{inverse roles} if the relativizer is dependent on a verb. 
The AMR for \textit{A boy who believes}, e.g.,
contains an inverse ARG0 role.
Other types of relative clauses, \textit{Noun Compound Expansions}, reveal a semantic relation 
between compound nouns. Such expansions can be expressed in various ways:

{\small
\eenumsentence{
\item \textit{A man is playing a \colorbox{lavender(web)}{flute made of bamboo}}\vspace*{-3ex}
\item \textit{A man is playing a  \colorbox{lavender(web)}{bamboo flute}}
}

\eenumsentence{
\item \textit{A child is running in and out of the \colorbox{lavender(web)}{waves of the ocean}}\vspace*{-3ex}
\item \textit{A child is running in and out of the \colorbox{lavender(web)}{ocean waves}}
}}

While the expansions in (\ex{-1}a, \ex{0}a) differ (\textit{made of} vs.\ \textit{of}), the two compound nouns in  (\ex{-1}b) and (\ex{0}b)
are connected with same AMR relation
\texttt{:part-of}, which 
reveals their semantic relation. The expansion in (\ex{-1}a), by contrast, 
emphasizes the process of the flute being \textit{made}, which is reflected in its AMR (see Fig.\ \ref{fig:sub2}, \ref{A:sub1}).
Hence, whenever we compare sentences that make use of a noun compound or an expansion of it, they may differ in their textual \textit{and} their AMR representations, which can have implications for different types of metrics. 

\subsection{Lexical Phenomena}

\textbf{Articles}
AMR does not specify articles, so
the sentence variants \{\underline{\textit{A}$\mid$\textit{The}}\} \textit{child is playing.} yield identical AMRs.
I.e., it cannot distinguish sentences differing in definiteness of an article. Our \textit{CheckList} includes pairs exhibiting such differences.

\textbf{Antonymy}
denotes a relation of contrast that can apply to
\textit{adjectives}, \textit{adverbs}, \textit{nouns}, \textit{prepositions} or \textit{verbs}. 
In AMR, antonymy is either implicit for concept pairs
or represented by negating a concept with \texttt{:polarity -} ~(Fig.\ \ref{fig:ant1} in \ref{A:antonymy}).

Note that human ratings in STS and SICK differ for  antonymy and negation. While in STS, antonymy and negation are penalized with low similarity scores, this is different
for SICK, which rates \textit{semantic relatedness} of sentences. Pairs including a single opposing concept 
may yield higher scores than comparison to a random sentence. This must be observed when interpreting \textit{CheckList} results.

\textbf{Hypernymy and Hyponymy,} and the 
derived \textbf{Co-Hyponymy} relation, while known from WordNet, are not explicitly
expressed between AMR concepts. They form the basis for inferential relations between sentences and  
play an important role in judging NLG quality from a semantic view. Often, a candidate may differ from its reference sentence by resorting to a superordinate, less specific concept, but may combine it with a differentiating modifier, yielding an equivalent meaning. Equivalence of compositional meaning is difficult to capture for word-based and lexical NLG metrics, and is even more challenging for metrics based on structured meaning representations. \textbf{Co-Hyponymy}, however, involves contrast and interferes with \textbf{Antonymy} and \textbf{Negation}.

\textbf{(Partial) Synonymy} 
We distinguish \textit{total} and \textit{partial} synonymy. In the former, linguistic expressions are interchangeable without restriction, while in the latter  this may hold in a context given their denotative 
meaning, may not hold when considering their
connotative meaning \citep{edmonds-hirst-2002-near}. Examples are \textit{lie -- untruth}, or \textit{task -- job}. While the former type is unproblematic for meaning-oriented, lexical NLG metrics, the latter is not, as it requires judging contextual conditions.
Since AMR specifies abstract concepts, choosing contextually adequate synonyms is a challenge, and contextualized  metrics may have an advantage.

\section{Interpreted Evaluation of NLG Metrics}
\label{sec:evaluation}

\subsection{Datasets and Statistics}

We sampled 939 sentence pairs, each differing  in a single phenomenon from SICK (877) and STS (62)\footnote{Distributions of phenomena and human scores 
in \ref{A:statistics}.}, parsed them into AMRs using the parser of \citet{raffel_t5} and manually corrected them.\footnote{Manual correction was performed by two of the authors.}

\textbf{STS (Semantic Textual Similarity).}
Since the first SemEval STS task 
\citep{agirre-etal-2012-semeval},
a total of 15,459 sentence pairs were created in follow-up challenges. 
Each sentence pair is annotated for semantic similarity on a 
Likert
scale from 5: "completely equivalent" to 0: “on different topics”.

\textbf{SICK: Sentences Involving Compositional Knowledge} by \citet{marelli-etal-2014-sick} contains 10,000 English sentence pairs, annotated for \textit{semantic relatedness} and \textit{entailment}.
Pairs were normalized, expanded using specific linguistic phenomena, and finally paired with one another. Due to this 
process, pairs often differ by  single linguistic phenomena, making them well suited for our aims.
The sentence pairs were rated for semantic relatedness on a five-point Likert scale, from 1: “completely unrelated” to 5: “very related”. 

Since the annotations on SICK and STS are not equivalent, they will  be analyzed separately. 

\subsection{Experimental Setup}

\textbf{Metrics} All metrics except GraCo  use existing implementations. To enhance comparability between metrics, we standardize and normalize the scores of every metric and the annotated human scores (see \ref{A:hyperparameters} for details on both).

\begin{table*}[t]
\resizebox{\linewidth}{!}{
\begin{tabular}{lllllllllll|l}
\hline
                  & Antonymy      & Article      & Co-Hyp.   & Hyponymy      & Negation     & Omission     & Part.\ Syn.ymy   & Passive       & Sem. Roles   & Sub. Clauses   & Overall      \\
\hline
 Ann. Score       & 0.614        & 0.977        & 0.628         & 0.863        & 0.597        & 0.86         & 0.941              & 0.976        & 0.6              & 0.963                 & 0.789        \\\hline
 BLEU             & 0.672 ± \underline{\textit{0.19}} & 0.772 ± 0.21 & 0.775 ± 0.22  & 0.72 ± 0.18  & 0.582 ± 0.2  & 0.645 ± 0.23 & 0.734 ± 0.22       & 0.108 ± 0.87 & 0.298 ± 0.3      & 0.579 ± 0.38          & 0.611 ± 0.28 \\
 chrF++           & 0.796 ± 0.2  & 0.865 ± 0.11 & 0.794 ± 0.2   & 0.779 ± 0.12 & 0.846 ± 0.25 & 0.728 ± 0.14 & 0.798 ± 0.15       & 0.339 ± 0.64 & 0.669 ± 0.12     & 0.733 ± 0.23          & 0.75 ± 0.22  \\
 Meteor           & 0.421 ± 0.24 & 0.605 ± 0.37 & 0.444 ± 0.22  & 0.669 ± 0.26 & 0.46 ± \underline{\textit{0.16}}  & 0.466 ± 0.39 & 0.808 ± 0.18       & 0.258 ± 0.72 & 0.415 ± 0.19     & 0.408 ± 0.56          & 0.482 ± 0.33 \\
 \textsc{BERTS}core       & 0.868 ± 0.26 & 0.953 ± 0.04 & 0.854 ± 0.24  & 0.86 ± \underline{\textit{0.08}}  & 0.749 ± 0.17 & 0.813 ± \underline{\textit{0.08}} & 0.925 ± \textbf{0.04}       & 0.512 ± 0.46 & 0.726 ± 0.16     & 0.783 ± 0.18          & 0.805 ± 0.17 \\
 \textsc{Smatch}           & 0.793 ± 0.22 & 0.998 ± \textbf{0.02} & 0.833 ± 0.22  & 0.83 ± \textbf{0.07}  & 0.921 ± 0.32 & 0.844 ± \textbf{0.06} & 0.829 ± 0.12       & 0.995 ± \textbf{0.03} & 0.647 ± \underline{\textit{0.11}}     & 0.917 ± 0.09          & 0.877 ± \textbf{0.14} \\
 \textsc{S$^2$match}          & 0.793 ± 0.22 & 0.998 ± \textbf{0.02} & 0.838 ± 0.23  & 0.831 ± \textbf{0.07} & 0.921 ± 0.32 & 0.844 ± \textbf{0.06} & 0.829 ± 0.12       & 0.995 ± \textbf{0.03} & 0.647 ± \underline{\textit{0.11}}     & 0.917 ± 0.09        & 0.877 ± \textbf{0.14} \\
 WLK              & 0.575 ± \textbf{0.16} & 0.989 ± \underline{\textit{0.03}} & 0.586 ± \textbf{0.16}  & 0.539 ± 0.32 & 0.791 ± 0.2  & 0.782 ± 0.1  & 0.614 ± 0.33       & 0.993 ± \textbf{0.03} & 0.525 ± \textbf{0.1}      & 0.896 ± 0.11          & 0.745 ± \underline{\textit{0.16}} \\
 WWLK             & 0.76 ± 0.21  & 0.996 ± \underline{\textit{0.03}} & 0.736 ± \underline{\textit{0.19}}  & 0.721 ± 0.16 & 0.644 ± \textbf{0.15} & 0.685 ± 0.18 & 0.734 ± 0.21       & 0.994 ± \textbf{0.03} & 0.936 ± 0.34     & 0.907 ± 0.1           & 0.774 ± \textbf{0.14} \\
\hline
 \textsc{GraCo}$_{gl}$        & 0.952 ± 0.36 & 1.0 ± \textbf{0.02}   & 0.97 ± 0.34   & 0.963 ± 0.11 & 0.974 ± 0.38 & 0.926 ± 0.13 & 0.975 ± \underline{\textit{0.05}}       & 0.936 ± 0.06 & 0.998 ± 0.4      & 0.992 ± \textbf{0.03}          & 0.961 ± 0.2  \\
 \textsc{GraCo}$_{gl}^{red}$ & 0.883 ± 0.35 & 1.0 ± \textbf{0.02}   & 0.942 ± 0.32  & 0.933 ± 0.09 & 0.381 ± 0.23 & 0.277 ± 0.59 & 0.951 ± \underline{\textit{0.05}}       & 0.93 ± 0.06  & 1.0 ± 0.4        & 0.853 ± 0.16          & 0.711 ± 0.26 \\
 \textsc{GraCo}              & 0.952 ± 0.34 & 0.969 ± 0.04 & 0.959 ± 0.33  & 0.949 ± 0.11 & 0.942 ± 0.35 & 0.935 ± 0.11 & 0.965 ± \underline{\textit{0.05}}       & 0.938 ± \underline{\textit{0.05}} & 0.985 ± 0.38     & 0.946 ± \underline{\textit{0.04}}          & 0.948 ± 0.19 \\
 \textsc{GraCo}$^{red}$       & 0.875 ± 0.32 & 1.0 ± \textbf{0.02}   & 0.91 ± 0.29   & 0.915 ± 0.11 & 0.497 ± 0.24 & 0.447 ± 0.43 & 0.937 ± 0.06       & 0.92 ± 0.07  & 0.92 ± 0.39      & 0.865 ± 0.14          & 0.755 ± 0.23 \\
\hline
\end{tabular}}
\caption{Avg.\ normalized score
\& mean abs.\ deviation (most indicative, lower is better) from human score for SICK.
}
\label{tab:av_sick}

\end{table*}

\begin{table}[h]
\resizebox{\linewidth}{!}{%
\begin{tabular}{@{}lllllllllllll@{}}
\hline
                 &   Ant.my &   Art. &   CoHyp &   Hyp &   Neg &   Omiss &   P.Syn &   Pass &   SRL &   
                 Sb.Cl &   Ovll\\
\hline
BLEU             &     0.492 &     0.34  &         \underline{\textit{0.54}}  &     0.419 &      0.433 &      0.459 &              0.391 &    \underline{\textit{0.335}} &            0.469 &                 0.321 &     0.424\\
chrF++           &     0.5   &     0.342 &         0.523 &     0.437 &      0.441 &      \underline{\textit{0.489}} &              0.435 &    0.303 &            \underline{\textit{0.562}} &                 0.336 &     0.367\\
Meteor           &     \textbf{0.538} &     0.35  &         \textbf{0.564} &     \textbf{0.494} &      0.441 &      0.435 &              \textbf{0.524} &    0.322 &            0.438 &                 0.365 &     0.463\\
\textsc{BERTS}c       &     0.483 &     0.36  &         0.505 &     \underline{\textit{0.469}} &      0.473 &      \textbf{0.523} &              0.435 &    0.31  &            0.406 &                 0.355 &     0.47\\
\textsc{Smatch}           &     0.485 &     0.357 &         0.486 &     0.402 &      0.408 &      0.456 &              0.399 &    \textbf{0.349} &            0.406 &                 0.364 &     0.579\\
\textsc{S$^2$match}          &     0.484 &     0.357 &         0.474 &     0.395 &      0.408 &      0.456 &              0.399 &    \textbf{0.349} &            0.406 &                 0.364 &     0.578\\
WLK              &     \underline{\textit{0.516}} &     \underline{\textit{0.375}} &         0.509 &     0.413 &      0.429 &      0.471 &              0.349 &    \textbf{0.349} &            0.469 &                 0.363 &     \underline{\textit{0.628}}\\
WWLK             &     0.485 &     0.357 &         0.456 &     0.439 &      0.449 &      0.47  &              0.396 &    \textbf{0.349} &            0.469 &                 0.357 &     \textbf{0.636}\\
 \hline
\textsc{GraCo}$_{glo}$         &     0.489 &     \textbf{0.385} &         0.469 &     0.436 &      0.458 &      0.415 &              0.296 &    0.302 &            0.219 &                 0.368 &     0.511\\
GraCo$_{glo}^{red}$ &     0.437 &     0.367 &         0.509 &     0.406 &      \underline{\textit{0.496}} &      0.405 &              0.402 &    0.305 &            0.188 &                 0.378 &     0.553\\
\textsc{GraCo}                &     0.473 &     0.292 &         0.497 &     0.411 &      0.428 &      0.46  &              \underline{\textit{0.485}} &    0.321 &            \textbf{0.625} &                 \textbf{0.46}  &     0.449\\
GraCo$^{red}$ &     0.433 &     0.367 &         0.481 &     0.416 &      \textbf{0.505} &      0.418 &              0.444 &    0.327 &            0.219 &                 \underline{\textit{0.384}} &     0.565\\
\hline
\end{tabular}}
    \caption{Pairwise ranking scores for the SICK test cases.}
    \label{tab:rank_sick}

\end{table}

\textbf{Evaluation metrics for metric performance} We compute
\textbf{i) Correlations} of the metric scores with the human scores using \textit{Spearman's rho}. 
\textbf{ii) Pairwise Ranking scores} 
for all metrics, 
where for each phenomenon we consider all possible combinations of pairs $(x,y)$ and $(x',y')$.
A metric $m$ scores one point if the relation between the predicted scores $m(x, y)$ and $m(x', y')$ for the given pairs corresponds to the relation between their
human scores $h(x, y)$ and $h(x', y')$.
If for instance $h(x,y) < h(x', y')$, metric $m$ earns one point if 

\vspace*{-6mm}
\begin{align*}
    & m(x, y) < m(x', y')\quad\land\\ 
     |&m(x, y) - m(x', y')| > \tau
\end{align*}
\vspace*{-6mm}

 where $\tau$ is a threshold we define
 as the fifth percentile of all scores. We define
 $m(x, y)$ = $m(x', y')$ if $|m(x, y) - m(x', y')| \leq \tau$.
\textbf{iii)} Mean Average score and its \textbf{Mean Absolute Deviation} (MAD) from the human score over test cases.

\subsection{Hypotheses}
\label{sec:hypotheses}
  
We state hypotheses on how various metrics are expected to perform for selected phenomena.\footnote{Due to space restrictions, 
we only discuss a selection, which we mark with \textcolor{green}{\cmark}Hx vs.\ \textcolor{red}{\xmark}Hx if (un)supported by results.}
    
\textbf{H1: \textit{gM} vs.\ \textit{tM}} AMR metrics are less sensitive to surface variation than textual metrics. This can be beneficial when variations have a mild impact on human judgements of similarity
(\textit{Passive}, \textit{Articles}), but may have adverse effects when the impact is high. This may happen with  
\textit{Antonymy}, if the metric cannot capture relevant differences in lexical meaning, as in \textsc{Smatch}.

We expect BERTScore to compete with \textit{gM} metrics, due to its contextualized representations. 
In general we expect all AMR metrics to have an advantage over textual metrics, except for \textsc{BERTS}core, in detecting \textit{Switched Roles}, since they explicitly represent argument roles. 

\textbf{H2: Impact of small substrings or subgraphs} 
Irrespective of differences in human judgement for \textit{Anto\-nymy, Co-hyponymy} and \textit{Negation} between SICK vs.\ STS (cf.\ \S 4), metrics can differ in how strongly a contrast at token or concept level affects a pair's overall rating.
In such cases only few triples may differ between sentence pairs, so we don't expect \textsc{S($^2$)match} to reflect  strong drops in human score. \textsc{W(W)lk} may fare better, as its kernel can capture a wider context of a given node.
\textsc{BERTS}core faces similar problems when small text portions cause a strong contrast, but its contextualization
may reflect the impact of neighboring words, an effect that could be shared with \textsc{W(W)lk}. 

While all prior metrics
compute scores over the entire sentences, 
\textsc{GraCo}$^{red}$ only considers local subgraphs restricted to \textit{differing} nodes. We expect this to be beneficial for phenomena like \textit{Negation}.

\textbf{H3: Capturing (dis)similarity}
We expect \textsc{S$^{2}$Match} and \textsc{W(W)lk} to perform closer to human judgement than \textsc{Smatch} for sentences that differ by semantically similar or closely related words, e.g., with \textit{Partial Synonymy} or \textit{Hyponymy}. 
The same 
should hold true for Meteor as opposed to BLEU and chrF++, since it accounts for synonyms and paraphrases.
\textsc{W(W)lk} is expected to capture compositional similarity (\textit{young cat -- kitten}) 
better than \textsc{S$^{2}$match}, which is purely lexical. 
But 
\textsc{S$^2$match} and \textsc{WWlk} could perform worse for \textit{Antonymy}, since antonyms 
tend to be close to each other in latent space
\citep{antonyms_similarity_samenko}.

\subsection{Results and Analyses}
\label{sec:results_analysis}
Results are displayed in Tables 3 and 4 for SICK.\footnote{STS results are seen in Tables \ref{tab:av_sts}, \ref{tab:rank_sts} and Fig.\ \ref{fig:sts_corr}, in \ref{A:STS_res}.} Fig.\ \ref{fig:sick_corr} displays an aggregated view of correlations between the metric scores and human scores for individual phenomena. Finally, Table \ref{tab:summarization_sick} presents a summary for all metrics and the phenomena they perform best or 2nd best on, according to our three evaluation metrics: ranking score, MAD and correlation to human judgement scores.

The \textit{gM} metrics \textsc{W(W)LK} show best overall performance, sharing 1st place with \textsc{S($^2$)Match} in SICK and obtaining first place in pairwise ranking, and we see  top places being achieved for 4-5 phenomena (\textcolor{green}{\cmark} H1, \textcolor{green}{\cmark} H3). But \textsc{S($^2$)Match} produce very similar scores across the board (\textcolor{red}{\xmark} H3).

Among symbolic \textit{tM} metrics, \textit{Meteor} performs best in ranking score, and \textit{chrF++} for MAD. 
\textsc{BERTS}core performs better than symbolic \textit{tM} metrics overall, except for ranking score for STS, where it only fails on \textit{Aspect} (\textcolor{green}{\cmark}H1).
But it falls behind \textit{gM} and most \textit{hyM} metrics in \textit{overall} scores.\\
\textsc{GraCo} performance varies across phenomena and its variants. It occupies 1st and 2nd places in ranking score for \textit{Neg} in SICK in the \textit{reduced} variant, where the drop in avg score and MAD is striking (\textcolor{green}{\cmark}H2). For other phenomena, the performance aligns with the other \textit{gM} metrics. This suggests that the connectivity score captures most lexical phenomena well -- while for \textit{SRL} this is evidently not sufficient (\textcolor{green}{\cmark}H1).

Beyond tendencies in overall results, 
we now focus on observations for single phenomena.

\begin{figure}[t!]
 \input{figures/sick_corr.tex}
 \end{figure}

\begin{table*}[t]
\centering
\resizebox{0.9\linewidth}{!}{%
\begin{tabular}{l|lll}
                 & \textbf{Best} \& \underline{2nd Best} Ranking Scores & \textbf{Best} \& \underline{2nd Best} MAD & \textbf{Highest} \& \underline{2nd Highest} Correlation w/ Human\\
BLEU     &  \underline{Passive}, \underline{Co-Hyp.} & \underline{Antonymy} & \underline{Co-Hyp.}, \underline{SRL}\\
chrF++           &  \underline{Omission}, \underline{SRL} &  & \underline{Omission}\\
Meteor          &  \textbf{Co-Hyp.}, \textbf{Antonymy}, \textbf{Part. Synonymy}, \textbf{Hyp.} & \underline{Negation} & \textbf{Part. Synonymy}, \textbf{Antonymy}, \textbf{Co-Hyp.}, \textbf{Hyp.}\\
\textsc{BERTS}c     &  \textbf{Omission}, \underline{Hyp.} & \textbf{Part. Synonymy}, \underline{Omission}, \underline{Hyp.} & \textbf{Omission}, \underline{Hyp.}\\
\midrule
\textsc{Smatch}         &  \textbf{Passive} & \textbf{Article}, \textbf{Passive}, \textbf{Omission}, \textbf{Hyp.}, \underline{SRL} & \textbf{Passive}\\
\textsc{S$^2$match}      &  \textbf{Passive} & \textbf{Article}, \textbf{Passive}, \textbf{Omission}, \textbf{Hyp.}, \underline{SRL} & \textbf{Passive}\\
WLK              &  \textbf{Passive}, \underline{Article}, \underline{Antonymy} & \textbf{Passive}, \textbf{SRL}, \textbf{Antonymy}, \textbf{Co-Hyp.}, \underline{Article} & \textbf{Passive}, \underline{Antonymy}\\
WWLK        &  \textbf{Passive} & \textbf{Passive}, \textbf{Negation}, \underline{Article}, \underline{Co-Hyp.} & \textbf{Passive}\\
\midrule
\textsc{GraCo}$_{glo}$ &  \textbf{Article} & \textbf{Article}, \textbf{Sub. Clauses}, \underline{Part. Synonymy} & \textbf{Article}\\
GraCo$_{glo}^{red}$  & \underline{Negation} & \textbf{Article}, \underline{Part. Synonymy} & \underline{Negation}\\
\textsc{GraCo}  &  \textbf{SRL}, \textbf{Sub. Clause}, \underline{Part. Synonymy} & \underline{Sub. Clauses}, \underline{Part. Synonymy}, \underline{Passive} & \textbf{SRL}, \textbf{Sub. Clauses}, \underline{Part. Synonymy}\\
GraCo$^{red}$  &  \textbf{Negation}, \underline{Sub. Clause} & \textbf{Article} & \textbf{Negation}, \underline{Sub. Clauses}
\end{tabular}}
    \caption{\textbf{Best} \& \underline{2$^{nd}$ Best} Metric Performances in Ranking Score, MAD, Corr.\ with Human Scores for SICK dataset.}
    \label{tab:summarization_sick}
\end{table*}
\begin{table*}[t]
\centering
\resizebox{0.8\linewidth}{!}{%
\begin{tabular}{@{}clccccccccccc@{}}
\toprule
&&\multicolumn{5}{c}{textual level}&\multicolumn{4}{c}{graph level} \\Type   & Metric& words & chars/pieces & lexicon & dense& contextual & concepts & sem. edges & sim. edges & dense & contextual\\
\midrule
       &  BLEU  & + & - & - & - & +\\
  $tM$ &  chrF++ & + & + & - & - & +\\
       & Meteor & + & - & + & - & -\\
       & BERTScore& - & + & - & + & +& & & &  \\
\midrule
      & \textsc{SMatch}    &   & & & & & + & + & -& - & - \\
  $gM$& \textsc{S$^{2}$Match} & & & & & & + & + & - & + & - \\ 
      & \textsc{Wlk} & & & & & & + & + & - & - & + \\
      & \textsc{WWlk}& & & & & & + & + & - & + & + \\
\midrule
 & \textsc{GraCo$_{glo}$}& + & -  & - & -& - & + & - & + & + & -\\
 $hyM$      & \textsc{GraCo$_{glo}^{red}$}& + & - & - & -& - & + & - & + & + & -\\
       & \textsc{GraCo}& + & - & - & -& -  & + & - & + & + & +\\
       & \textsc{GraCo$^{red}$}& + & - & - & -& - & + & - & + & + & +\\
\bottomrule
\end{tabular}%
}
\caption{Characterization of the used textual (\textit{tM}), graph-based (\textit{gM}) and hybrid (\textit{hyM}) metrics in terms of textual and graph-level properties. \textbf{textual level}: word/char/lexicon-based; \textbf{graph-level}: semantic vs. similarity edges;
\textbf{both levels}: dense = embedding-based representation; contextual = contextualized representation.
}
\label{tab:metrics}
\end{table*}

While  \textit{gM} generally outperform \textit{tM} metrics,
this doesn't necessarily hold for Meteor: it outperforms \textit{gM}  for phenomena 
reflecting lexical-semantic relations for SICK 
(Table \ref{tab:rank_sick}, Fig.\ \ref{fig:sick_corr}). The spike in correlation for \textit{Part.\ Syn.}\ is expected, as Meteor accounts for synonyms and paraphrases (\textcolor{green}{\cmark}H3). This may also explain its superior performance for \textit{(Co-)Hyponymy}.
 But its high performance for 
 \textit{Antonymy} is surprising (\textcolor{red}{\xmark}H3).

\textsc{S$^2$match} performing
very similar to \textsc{Smatch}
is most likely due to a high
threshold for allowing a soft match.
\textsc{GraCo} was
designed
to better represent semantic contrast
between sentences and their AMR graphs. We can see this 
reflected in a large drop of MAD for 
\textsc{GraCo}$^{red}$ in \textit{Negation}. 
In comparison, for \textit{Antonymy} we only see a relatively small drop in 
MAD.
This is because, for \textit{Negation}, \textsc{GraCo}$^{red}$
produces a bigger contrast between the connectivity scores as one of them is 1 for the empty graph.
For \textit{Antonymy} the scores are closer, 
since both graphs have neighbors. Another factor could be the proximity of antonyms in embedding space, which suggests that 
a threshold, similar to
\textsc{S$^2$match}, could be beneficial.

We also observe that \textsc{GraCo} using \textsc{BERT} outperforms \textsc{GraCo}$_{glo}$ in \textit{Part.Syn, SRL, SubCl} (Table \ref{tab:rank_sick},  
 Fig.\ \ref{fig:sick_corr}). This is unexpected since neither of them uses AMR relations. 
 This could be explained by the contextualized node embeddings that see context at textual level--combined with connectivity graphs that look at the sentence only via AMR nodes.
 
 Overall we see surprising effects with \textsc{GraCo}:
i) by restricting connectivity to local subgraphs for contrasting elements, it yields strong performance for \textit{Negation}; ii) it only focuses on AMR nodes, but 
the contrast with \textsc{GraCo}$_{glo}$ suggests that the contextualization helps to assess surface differences underlying \textit{SRL} and \textit{SubCl}.
The insights from \textsc{GraCo} could trigger ideas for improving a \textit{tM} metric like \textsc{BERTS}core, by computing it under a similar AMR lens, and handling \textit{Negation} in similar ways. It also suggests studying the use of \textsc{BERT} embeddings in \textsc{WWlk}, and seeking ways of integrating a comparable mechanism for \textit{Negation}.
As for \textit{tM} metrics, it came as a surprise to find Meteor keep 1st rank for lexical relations ((Co-)Hyp; (Partial)Syn, Antonymy), beyond \textsc{BERTS}core. 

\section{Conclusion}

We introduced an extensible \textit{CheckList} for mea\-ning-\-oriented NLG metrics that allows for comparison of a wide range of NLG metrics. 
It is interpreted by way of
offering test cases grouped by linguistic phenomena. 
Our analyses showcase how \textit{CheckList} can be used to compare metrics,
to reveal their strengths and weaknesses. They align
with a number of hypotheses, but also show surprising effects, opening avenues to further improve NLG evaluation metrics.
We propose a novel, hybrid similarity metric \textsc{GraCo} that builds cohesion graphs over contextualized AMR concept nodes. The metric can focus on contrastive subgraphs, which yields strong correlation with human judgements for negation. With regard to current practice in AMR-to-text evaluation, we find evidence that meaning-oriented graph-based metrics present advantages over typical  text-based metrics,  confirming the findings of \citet{opitz-frank-2021-towards,manning-etal-2020-human}. 
 Therefore we recommend to include graph metrics or hybrid graph- and textual metrics into AMR-to-text evaluation protocols. 
Our data and code will be publicly available.\footnote{\url{https://github.com/Heidelberg-NLP/NLG-CHECKLIST}} We welcome contributions 
to grow it.

\section*{Acknowledgements}
We thank the anonymous reviewers for useful feedback and suggestions.


\bibliography{anthology,custom}

\begin{thebibliography}{42}
\expandafter\ifx\csname natexlab\endcsname\relax\def\natexlab#1{#1}\fi

\bibitem[{Agirre et~al.(2012)Agirre, Cer, Diab, and
  Gonzalez-Agirre}]{agirre-etal-2012-semeval}
Eneko Agirre, Daniel Cer, Mona Diab, and Aitor Gonzalez-Agirre. 2012.
\newblock \href {https://aclanthology.org/S12-1051} {{S}em{E}val-2012 task 6: A
  pilot on semantic textual similarity}.
\newblock In \emph{*{SEM} 2012: The First Joint Conference on Lexical and
  Computational Semantics {--} Volume 1: Proceedings of the main conference and
  the shared task, and Volume 2: Proceedings of the Sixth International
  Workshop on Semantic Evaluation ({S}em{E}val 2012)}, pages 385--393,
  Montr{\'e}al, Canada. Association for Computational Linguistics.

\bibitem[{Banarescu et~al.(2013)Banarescu, Bonial, Cai, Georgescu, Griffitt,
  Hermjakob, Knight, Koehn, Palmer, and
  Schneider}]{banarescu-etal-2013-abstract}
Laura Banarescu, Claire Bonial, Shu Cai, Madalina Georgescu, Kira Griffitt, Ulf
  Hermjakob, Kevin Knight, Philipp Koehn, Martha Palmer, and Nathan Schneider.
  2013.
\newblock \href {https://aclanthology.org/W13-2322} {{A}bstract {M}eaning
  {R}epresentation for sembanking}.
\newblock In \emph{Proceedings of the 7th Linguistic Annotation Workshop and
  Interoperability with Discourse}, pages 178--186, Sofia, Bulgaria.
  Association for Computational Linguistics.

\bibitem[{Banarescu et~al.(2019)Banarescu, Bonial, Cai, Georgescu, Griffitt,
  Hermjakob, Knight, Koehn, Palmer, and Schneider}]{amr_guidelines}
Laura Banarescu, Claire Bonial, Shu Cai, Madalina Georgescu, Kira Griffitt, Ulf
  Hermjakob, Kevin Knight, Philipp Koehn, Martha Palmer, and Nathan Schneider.
  2019.
\newblock \href {https://github.com/amrisi/amr-guidelines/blob/master/amr.md}
  {Abstract meaning representation (amr) 1.2.6 specification}.

\bibitem[{Bird et~al.(2009)Bird, Klein, and Loper}]{nltk}
Steven Bird, Ewan Klein, and Edward Loper. 2009.
\newblock \emph{Natural language processing with Python: analyzing text with
  the natural language toolkit}.
\newblock " O'Reilly Media, Inc.".

\bibitem[{Blloshmi et~al.(2021)Blloshmi, Bevilacqua, Fabiano, Caruso, and
  Navigli}]{blloshmi-etal-2021-spring}
Rexhina Blloshmi, Michele Bevilacqua, Edoardo Fabiano, Valentina Caruso, and
  Roberto Navigli. 2021.
\newblock \href {https://doi.org/10.18653/v1/2021.emnlp-demo.16} {{SPRING}
  {G}oes {O}nline: {E}nd-to-{E}nd {AMR} {P}arsing and {G}eneration}.
\newblock In \emph{Proceedings of the 2021 Conference on Empirical Methods in
  Natural Language Processing: System Demonstrations}, pages 134--142, Online
  and Punta Cana, Dominican Republic. Association for Computational
  Linguistics.

\bibitem[{Cai and Knight(2013)}]{cai-knight-2013-smatch}
Shu Cai and Kevin Knight. 2013.
\newblock \href {https://aclanthology.org/P13-2131} {{S}match: an evaluation
  metric for semantic feature structures}.
\newblock In \emph{Proceedings of the 51st Annual Meeting of the Association
  for Computational Linguistics (Volume 2: Short Papers)}, pages 748--752,
  Sofia, Bulgaria. Association for Computational Linguistics.

\bibitem[{Chandrasekaran and Mago(2021)}]{10.1145/3440755}
Dhivya Chandrasekaran and Vijay Mago. 2021.
\newblock \href {https://doi.org/10.1145/3440755} {Evolution of semantic
  similarity—a survey}.
\newblock \emph{ACM Comput. Surv.}, 54(2).

\bibitem[{Cilibrasi and Vitanyi(2007)}]{cilibrasi2007google}
Rudi~L Cilibrasi and Paul~MB Vitanyi. 2007.
\newblock The google similarity distance.
\newblock \emph{IEEE Transactions on knowledge and data engineering},
  19(3):370--383.

\bibitem[{Conneau and Kiela(2018)}]{conneau-kiela-2018-senteval}
Alexis Conneau and Douwe Kiela. 2018.
\newblock \href {https://aclanthology.org/L18-1269} {{S}ent{E}val: An
  evaluation toolkit for universal sentence representations}.
\newblock In \emph{Proceedings of the Eleventh International Conference on
  Language Resources and Evaluation ({LREC} 2018)}, Miyazaki, Japan. European
  Language Resources Association (ELRA).

\bibitem[{Conneau et~al.(2017)Conneau, Kiela, Schwenk, Barrault, and
  Bordes}]{conneau-etal-2017-supervised}
Alexis Conneau, Douwe Kiela, Holger Schwenk, Lo{\"\i}c Barrault, and Antoine
  Bordes. 2017.
\newblock \href {https://doi.org/10.18653/v1/D17-1070} {Supervised learning of
  universal sentence representations from natural language inference data}.
\newblock In \emph{Proceedings of the 2017 Conference on Empirical Methods in
  Natural Language Processing}, pages 670--680, Copenhagen, Denmark.
  Association for Computational Linguistics.

\bibitem[{Denkowski and Lavie(2014)}]{denkowski-lavie-2014-meteor}
Michael Denkowski and Alon Lavie. 2014.
\newblock \href {https://doi.org/10.3115/v1/W14-3348} {Meteor universal:
  Language specific translation evaluation for any target language}.
\newblock In \emph{Proceedings of the Ninth Workshop on Statistical Machine
  Translation}, pages 376--380, Baltimore, Maryland, USA. Association for
  Computational Linguistics.

\bibitem[{Devlin et~al.(2019)Devlin, Chang, Lee, and
  Toutanova}]{devlin-etal-2019-bert}
Jacob Devlin, Ming-Wei Chang, Kenton Lee, and Kristina Toutanova. 2019.
\newblock \href {https://doi.org/10.18653/v1/N19-1423} {{BERT}: Pre-training of
  deep bidirectional transformers for language understanding}.
\newblock In \emph{Proceedings of the 2019 Conference of the North {A}merican
  Chapter of the Association for Computational Linguistics: Human Language
  Technologies, Volume 1 (Long and Short Papers)}, pages 4171--4186,
  Minneapolis, Minnesota. Association for Computational Linguistics.

\bibitem[{Edmonds and Hirst(2002)}]{edmonds-hirst-2002-near}
Philip Edmonds and Graeme Hirst. 2002.
\newblock \href {https://doi.org/10.1162/089120102760173625} {Near-synonymy and
  lexical choice}.
\newblock \emph{Computational Linguistics}, 28(2):105--144.

\bibitem[{Flanigan et~al.(2014{\natexlab{a}})Flanigan, Thomson, Carbonell,
  Dyer, and Smith}]{flanigan-etal-2014-discriminative}
Jeffrey Flanigan, Sam Thomson, Jaime Carbonell, Chris Dyer, and Noah~A. Smith.
  2014{\natexlab{a}}.
\newblock \href {https://doi.org/10.3115/v1/P14-1134} {A discriminative
  graph-based parser for the {A}bstract {M}eaning {R}epresentation}.
\newblock In \emph{Proceedings of the 52nd Annual Meeting of the Association
  for Computational Linguistics (Volume 1: Long Papers)}, pages 1426--1436,
  Baltimore, Maryland. Association for Computational Linguistics.

\bibitem[{Flanigan et~al.(2014{\natexlab{b}})Flanigan, Thomson, Carbonell,
  Dyer, and Smith}]{jamr}
Jeffrey Flanigan, Sam Thomson, Jaime Carbonell, Chris Dyer, and Noah~A. Smith.
  2014{\natexlab{b}}.
\newblock \href {https://doi.org/10.3115/v1/P14-1134} {A discriminative
  graph-based parser for the {A}bstract {M}eaning {R}epresentation}.
\newblock In \emph{Proceedings of the 52nd Annual Meeting of the Association
  for Computational Linguistics (Volume 1: Long Papers)}, pages 1426--1436,
  Baltimore, Maryland. Association for Computational Linguistics.

\bibitem[{Haagsma et~al.(2018)Haagsma, Nissim, and
  Bos}]{haagsma-etal-2018-side}
Hessel Haagsma, Malvina Nissim, and Johan Bos. 2018.
\newblock \href {https://aclanthology.org/W18-4919} {The other side of the
  coin: Unsupervised disambiguation of potentially idiomatic expressions by
  contrasting senses}.
\newblock In \emph{Proceedings of the Joint Workshop on Linguistic Annotation,
  Multiword Expressions and Constructions ({LAW}-{MWE}-{C}x{G}-2018)}, pages
  178--184, Santa Fe, New Mexico, USA. Association for Computational
  Linguistics.

\bibitem[{Konstas et~al.(2017)Konstas, Iyer, Yatskar, Choi, and
  Zettlemoyer}]{konstas-etal-2017-neural}
Ioannis Konstas, Srinivasan Iyer, Mark Yatskar, Yejin Choi, and Luke
  Zettlemoyer. 2017.
\newblock \href {https://doi.org/10.18653/v1/P17-1014} {Neural {AMR}:
  Sequence-to-sequence models for parsing and generation}.
\newblock In \emph{Proceedings of the 55th Annual Meeting of the Association
  for Computational Linguistics (Volume 1: Long Papers)}, pages 146--157,
  Vancouver, Canada. Association for Computational Linguistics.

\bibitem[{Lavie and Agarwal(2007)}]{lavie-agarwal-2007-meteor}
Alon Lavie and Abhaya Agarwal. 2007.
\newblock \href {https://aclanthology.org/W07-0734} {{METEOR}: An automatic
  metric for {MT} evaluation with high levels of correlation with human
  judgments}.
\newblock In \emph{Proceedings of the Second Workshop on Statistical Machine
  Translation}, pages 228--231, Prague, Czech Republic. Association for
  Computational Linguistics.

\bibitem[{Lehmann et~al.(1996)Lehmann, Oepen, Regnier-Prost, Netter, Lux,
  Klein, Falkedal, Fouvry, Estival, Dauphin, Compagnion, Baur, Balkan, and
  Arnold}]{lehmann-etal-1996-tsnlp}
Sabine Lehmann, Stephan Oepen, Sylvie Regnier-Prost, Klaus Netter, Veronika
  Lux, Judith Klein, Kirsten Falkedal, Frederik Fouvry, Dominique Estival, Eva
  Dauphin, Herve Compagnion, Judith Baur, Lorna Balkan, and Doug Arnold. 1996.
\newblock \href {https://aclanthology.org/C96-2120} {{TSNLP} - test suites for
  natural language processing}.
\newblock In \emph{{COLING} 1996 Volume 2: The 16th International Conference on
  Computational Linguistics}.

\bibitem[{Manning et~al.(2020)Manning, Wein, and
  Schneider}]{manning-etal-2020-human}
Emma Manning, Shira Wein, and Nathan Schneider. 2020.
\newblock \href {https://doi.org/10.18653/v1/2020.coling-main.420} {A human
  evaluation of {AMR}-to-{E}nglish generation systems}.
\newblock In \emph{Proceedings of the 28th International Conference on
  Computational Linguistics}, pages 4773--4786, Barcelona, Spain (Online).
  International Committee on Computational Linguistics.

\bibitem[{Marelli et~al.(2014)Marelli, Menini, Baroni, Bentivogli, Bernardi,
  and Zamparelli}]{marelli-etal-2014-sick}
Marco Marelli, Stefano Menini, Marco Baroni, Luisa Bentivogli, Raffaella
  Bernardi, and Roberto Zamparelli. 2014.
\newblock \href
  {http://www.lrec-conf.org/proceedings/lrec2014/pdf/363_Paper.pdf} {A {SICK}
  cure for the evaluation of compositional distributional semantic models}.
\newblock In \emph{Proceedings of the Ninth International Conference on
  Language Resources and Evaluation ({LREC}'14)}, pages 216--223, Reykjavik,
  Iceland. European Language Resources Association (ELRA).

\bibitem[{May and Priyadarshi(2017)}]{may-priyadarshi-2017-semeval}
Jonathan May and Jay Priyadarshi. 2017.
\newblock \href {https://doi.org/10.18653/v1/S17-2090} {{S}em{E}val-2017 task
  9: {A}bstract {M}eaning {R}epresentation parsing and generation}.
\newblock In \emph{Proceedings of the 11th International Workshop on Semantic
  Evaluation ({S}em{E}val-2017)}, pages 536--545, Vancouver, Canada.
  Association for Computational Linguistics.

\bibitem[{Opitz et~al.(2021)Opitz, Daza, and
  Frank}]{opitz-etal-2021-weisfeiler}
Juri Opitz, Angel Daza, and Anette Frank. 2021.
\newblock \href {https://doi.org/10.1162/tacl_a_00435} {Weisfeiler-leman in the
  bamboo: Novel {AMR} graph metrics and a benchmark for {AMR} graph
  similarity}.
\newblock \emph{Transactions of the Association for Computational Linguistics},
  9:1425--1441.

\bibitem[{Opitz and Frank(2021)}]{opitz-frank-2021-towards}
Juri Opitz and Anette Frank. 2021.
\newblock \href {https://doi.org/10.18653/v1/2021.eacl-main.129} {Towards a
  decomposable metric for explainable evaluation of text generation from
  {AMR}}.
\newblock In \emph{Proceedings of the 16th Conference of the European Chapter
  of the Association for Computational Linguistics: Main Volume}, pages
  1504--1518, Online. Association for Computational Linguistics.

\bibitem[{Opitz et~al.(2020)Opitz, Parcalabescu, and
  Frank}]{opitz-etal-2020-amr}
Juri Opitz, Letitia Parcalabescu, and Anette Frank. 2020.
\newblock \href {https://doi.org/10.1162/tacl_a_00329} {{AMR} similarity
  metrics from principles}.
\newblock \emph{Transactions of the Association for Computational Linguistics},
  8:522--538.

\bibitem[{Papineni et~al.(2002)Papineni, Roukos, Ward, and
  Zhu}]{papineni-etal-2002-bleu}
Kishore Papineni, Salim Roukos, Todd Ward, and Wei-Jing Zhu. 2002.
\newblock \href {https://doi.org/10.3115/1073083.1073135} {{B}leu: a method for
  automatic evaluation of machine translation}.
\newblock In \emph{Proceedings of the 40th Annual Meeting of the Association
  for Computational Linguistics}, pages 311--318, Philadelphia, Pennsylvania,
  USA. Association for Computational Linguistics.

\bibitem[{Pennington et~al.(2014)Pennington, Socher, and
  Manning}]{pennington-etal-2014-glove}
Jeffrey Pennington, Richard Socher, and Christopher Manning. 2014.
\newblock \href {https://doi.org/10.3115/v1/D14-1162} {{G}lo{V}e: Global
  vectors for word representation}.
\newblock In \emph{Proceedings of the 2014 Conference on Empirical Methods in
  Natural Language Processing ({EMNLP})}, pages 1532--1543, Doha, Qatar.
  Association for Computational Linguistics.

\bibitem[{Popov(2017)}]{popov-2017-word}
Alexander Popov. 2017.
\newblock \href {https://doi.org/10.26615/issn.1314-9156.2017_004} {Word sense
  disambiguation with recurrent neural networks}.
\newblock In \emph{Proceedings of the Student Research Workshop Associated with
  {RANLP} 2017}, pages 25--34, Varna. INCOMA Ltd.

\bibitem[{Popovi{\'c}(2015)}]{popovic-2015-chrf}
Maja Popovi{\'c}. 2015.
\newblock \href {https://doi.org/10.18653/v1/W15-3049} {chr{F}: character
  n-gram {F}-score for automatic {MT} evaluation}.
\newblock In \emph{Proceedings of the Tenth Workshop on Statistical Machine
  Translation}, pages 392--395, Lisbon, Portugal. Association for Computational
  Linguistics.

\bibitem[{Popovi{\'c}(2016)}]{popovic-2016-chrf}
Maja Popovi{\'c}. 2016.
\newblock \href {https://doi.org/10.18653/v1/W16-2341} {chr{F} deconstructed:
  beta parameters and n-gram weights}.
\newblock In \emph{Proceedings of the First Conference on Machine Translation:
  Volume 2, Shared Task Papers}, pages 499--504, Berlin, Germany. Association
  for Computational Linguistics.

\bibitem[{Raffel et~al.(2019)Raffel, Shazeer, Roberts, Lee, Narang, Matena,
  Zhou, Li, and Liu}]{raffel_t5}
Colin Raffel, Noam Shazeer, Adam Roberts, Katherine Lee, Sharan Narang, Michael
  Matena, Yanqi Zhou, Wei Li, and Peter~J. Liu. 2019.
\newblock \href {http://arxiv.org/abs/1910.10683} {Exploring the limits of
  transfer learning with a unified text-to-text transformer}.
\newblock \emph{CoRR}, abs/1910.10683.

\bibitem[{Reimers and Gurevych(2019)}]{reimers-gurevych-2019-sentence}
Nils Reimers and Iryna Gurevych. 2019.
\newblock \href {https://doi.org/10.18653/v1/D19-1410} {Sentence-{BERT}:
  Sentence embeddings using {S}iamese {BERT}-networks}.
\newblock In \emph{Proceedings of the 2019 Conference on Empirical Methods in
  Natural Language Processing and the 9th International Joint Conference on
  Natural Language Processing (EMNLP-IJCNLP)}, pages 3982--3992, Hong Kong,
  China. Association for Computational Linguistics.

\bibitem[{Ribeiro et~al.(2020)Ribeiro, Wu, Guestrin, and
  Singh}]{ribeiro-etal-2020-beyond}
Marco~Tulio Ribeiro, Tongshuang Wu, Carlos Guestrin, and Sameer Singh. 2020.
\newblock \href {https://doi.org/10.18653/v1/2020.acl-main.442} {Beyond
  accuracy: Behavioral testing of {NLP} models with {C}heck{L}ist}.
\newblock In \emph{Proceedings of the 58th Annual Meeting of the Association
  for Computational Linguistics}, pages 4902--4912, Online. Association for
  Computational Linguistics.

\bibitem[{Samenko et~al.(2020)Samenko, Tikhonov, and
  Yamshchikov}]{antonyms_similarity_samenko}
Igor Samenko, Alexey Tikhonov, and Ivan~P. Yamshchikov. 2020.
\newblock \href {http://arxiv.org/abs/2004.12835} {Synonyms and antonyms:
  Embedded conflict}.
\newblock arXiv:2004.12835.

\bibitem[{Song et~al.(2018)Song, Zhang, Wang, and
  Gildea}]{song-etal-2018-graph}
Linfeng Song, Yue Zhang, Zhiguo Wang, and Daniel Gildea. 2018.
\newblock \href {https://doi.org/10.18653/v1/P18-1150} {A graph-to-sequence
  model for {AMR}-to-text generation}.
\newblock In \emph{Proceedings of the 56th Annual Meeting of the Association
  for Computational Linguistics (Volume 1: Long Papers)}, pages 1616--1626,
  Melbourne, Australia. Association for Computational Linguistics.

\bibitem[{Sporleder and Li(2009)}]{Sporleder-li-2009-unsupervised}
Caroline Sporleder and Linlin Li. 2009.
\newblock \href {https://aclanthology.org/E09-1086} {Unsupervised recognition
  of literal and non-literal use of idiomatic expressions}.
\newblock In \emph{Proceedings of the 12th Conference of the {E}uropean Chapter
  of the {ACL} ({EACL} 2009)}, pages 754--762, Athens, Greece. Association for
  Computational Linguistics.

\bibitem[{Stanojevi{\'c} et~al.(2015)Stanojevi{\'c}, Kamran, Koehn, and
  Bojar}]{stanojevic-etal-2015-results}
Milo{\v{s}} Stanojevi{\'c}, Amir Kamran, Philipp Koehn, and Ond{\v{r}}ej Bojar.
  2015.
\newblock \href {https://doi.org/10.18653/v1/W15-3031} {Results of the {WMT}15
  metrics shared task}.
\newblock In \emph{Proceedings of the Tenth Workshop on Statistical Machine
  Translation}, pages 256--273, Lisbon, Portugal. Association for Computational
  Linguistics.

\bibitem[{Wang et~al.(2020)Wang, Wan, and Jin}]{wang-etal-2020-amr}
Tianming Wang, Xiaojun Wan, and Hanqi Jin. 2020.
\newblock \href {https://doi.org/10.1162/tacl_a_00297} {{AMR}-to-text
  generation with graph transformer}.
\newblock \emph{Transactions of the Association for Computational Linguistics},
  8:19--33.

\bibitem[{Wu et~al.(2019)Wu, Ribeiro, Heer, and Weld}]{wu-etal-2019-errudite}
Tongshuang Wu, Marco~Tulio Ribeiro, Jeffrey Heer, and Daniel Weld. 2019.
\newblock \href {https://doi.org/10.18653/v1/P19-1073} {{E}rrudite: Scalable,
  reproducible, and testable error analysis}.
\newblock In \emph{Proceedings of the 57th Annual Meeting of the Association
  for Computational Linguistics}, pages 747--763, Florence, Italy. Association
  for Computational Linguistics.

\bibitem[{Xiao and Wang(2021)}]{xiao-wang-2021-hallucination}
Yijun Xiao and William~Yang Wang. 2021.
\newblock \href {https://doi.org/10.18653/v1/2021.eacl-main.236} {On
  hallucination and predictive uncertainty in conditional language generation}.
\newblock In \emph{Proceedings of the 16th Conference of the European Chapter
  of the Association for Computational Linguistics: Main Volume}, pages
  2734--2744, Online. Association for Computational Linguistics.

\bibitem[{Zhang et~al.(2020)Zhang, Kishore, Wu, Weinberger, and
  Artzi}]{BERTscore}
Tianyi Zhang, Varsha Kishore, Felix Wu, Kilian~Q. Weinberger, and Yoav Artzi.
  2020.
\newblock {BERTScore: Evaluating Text Generation with BERT}.
\newblock In \emph{Proceedings of the Eighth International Conference on
  Learning Representations (ICLR)}.

\bibitem[{Zhao et~al.(2019)Zhao, Peyrard, Liu, Gao, Meyer, and
  Eger}]{zhao-etal-2019-moverscore}
Wei Zhao, Maxime Peyrard, Fei Liu, Yang Gao, Christian~M. Meyer, and Steffen
  Eger. 2019.
\newblock \href {https://doi.org/10.18653/v1/D19-1053} {{M}over{S}core: Text
  generation evaluating with contextualized embeddings and earth mover
  distance}.
\newblock In \emph{Proceedings of the 2019 Conference on Empirical Methods in
  Natural Language Processing and the 9th International Joint Conference on
  Natural Language Processing (EMNLP-IJCNLP)}, pages 563--578, Hong Kong,
  China. Association for Computational Linguistics.

\end{thebibliography}
\bibliographystyle{acl_natbib}
\appendix

\section{Appendix}
\label{sec:appendix}

\subsection{\textit{CheckList}'s functionalities and resources} \label{checklist_descr}
As described in \S \ref{sec:aims}, \textit{CheckList} contains the selected sentence pairs as well as the corresponding AMR structures and their human score grouped by linguistic phenomena in \texttt{json} format. It further includes the assigned scores for the test instances as well as code to run the implementation for the following metrics:
\begin{itemize}
\setlength{\itemsep}{-1mm}
    \item BLEU
    \item Meteor
    \item chrF++
    \item \textsc{BERTS}core
    \item \textsc{Smatch}
    \item \textsc{S$^2$match}
    \item \textsc{WLK}
    \item \textsc{WWLK}
\end{itemize}

\textbf{Output.} The \textit{CheckList} can be run from the command line, printing an overview of the data used, accompanied by statistics concerning human judgement for each phenomenon. These statistics include the mean, median, standard deviation and standard error of the human scores. Finally, it will output tables displaying the overall results of the \textit{CheckList} (hereby, we use the evaluation measures that were also applied in the paper). If a metric were to be tested, it would furthermore print the correlation of that metric with the others in decreasing order.\\
The results for the phenomena are summarized in individual text files. These files once more list the statistics about the human score and then display the average scores of all metrics for that very phenomenon. Finally, each test case is listed, including the sentences as well as their AMR structures and the scores assigned to it by the metrics and the annotator.

\subsection{STS Results} \label{A:STS_res}
\begin{figure}
    \centering
    \includegraphics[width=8cm]{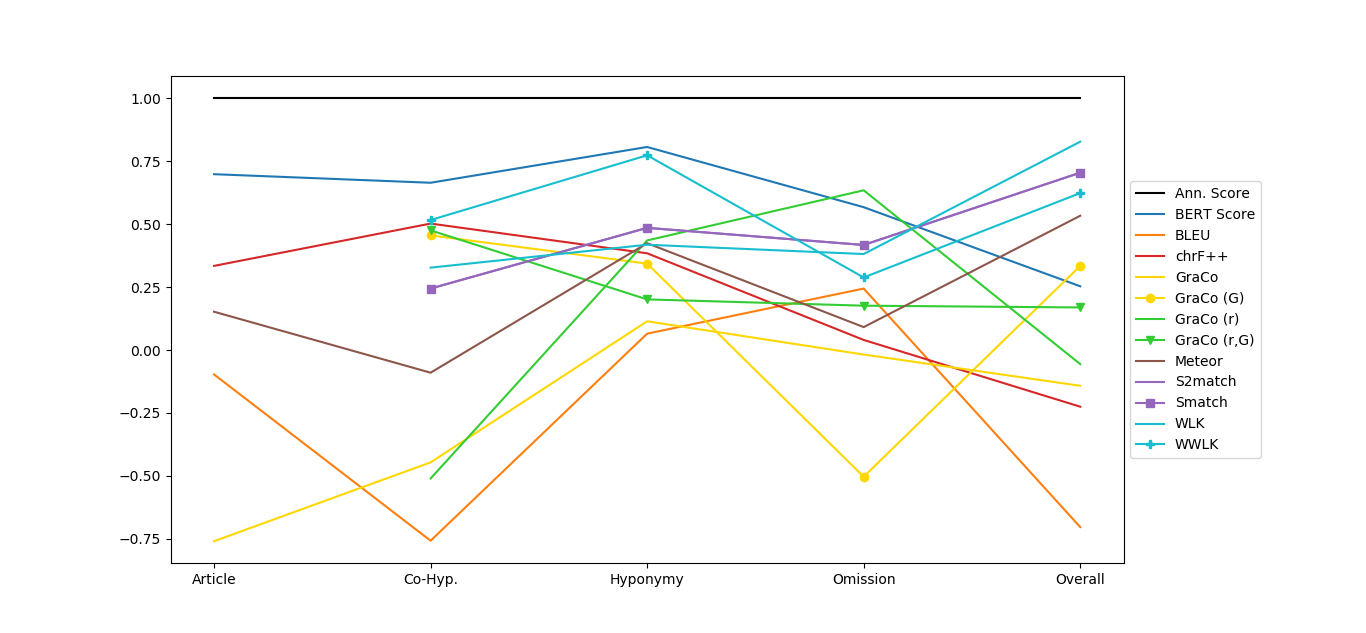}
    \caption{Spearman's rho correlation between metric and human scores for STS. \textit{Aspect} is not included since all annotated scores are 1.}
    \label{fig:sts_corr}
\end{figure}

Table \ref{tab:av_sts} and \ref{tab:rank_sts} and Fig. \ref{fig:sts_corr} demonstrate the results on the test cases selected from the STS data set. Table \ref{tab:metric_sts} shows a summary of metrics yielding Best and 2nd Best Results.

\begin{table}[hbt!]
\resizebox{\linewidth}{!}{
\begin{tabular}{@{}llllll|l@{}}
\hline
                  & Article      & Aspect       & Co-Hyponymy   & Hyponymy      & Omission     & Overall      \\
\hline
 Ann. Score       & 0.967        & 1.0          & 0.282         & 0.647        & 0.77         & 0.647        \\
 \hline
 BLEU             & 0.358 ± 0.61 & 0.155 ± 0.84 & 0.674 ± 0.48  & 0.58 ± 0.2   & 0.508 ± 0.27 & 0.503 ± 0.45 \\
 chrF++           & 0.661 ± 0.31 & 0.521 ± 0.48 & 0.661 ± 0.39  & 0.683 ± \underline{\textit{0.12}} & 0.707 ± 0.14 & 0.654 ± 0.29 \\
 Meteor           & 0.385 ± 0.58 & 0.557 ± 0.44 & 0.313 ± \textbf{0.2}   & 0.462 ± 0.3  & 0.407 ± 0.36 & 0.408 ± 0.33 \\
 \textsc{BERTS}core       & 0.863 ± 0.1  & 0.824 ± 0.18 & 0.838 ± 0.56  & 0.761 ± \underline{\textit{0.12}} & 0.801 ± \textbf{0.07} & 0.816 ± 0.26 \\
 \textsc{S$^2$match}          & 1.0 ± \textbf{0.03}   & 1.0 ± \textbf{0.0}    & 0.779 ± 0.5   & 0.737 ± 0.13 & 0.785 ± \underline{\textit{0.09}} & 0.83 ± 0.21  \\
 \textsc{Smatch}           & 1.0 ± \textbf{0.03}   & 1.0 ± \textbf{0.0}    & 0.779 ± 0.5   & 0.737 ± 0.13 & 0.785 ± \underline{\textit{0.09}} & 0.83 ± 0.21  \\
 WLK              & 1.0 ± \textbf{0.03}   & 1.0 ± \textbf{0.0}    & 0.459 ± \underline{\textit{0.25}}  & 0.426 ± 0.23 & 0.733 ± 0.11 & 0.659 ± \textbf{0.15} \\
 WWLK             & 1.0 ± \textbf{0.03}   & 1.0 ± \textbf{0.0}    & 0.689 ± 0.41  & 0.587 ± \textbf{0.1}  & 0.612 ± 0.19 & 0.732 ± \underline{\textit{0.2}}  \\
 \hline
 \textsc{GraCo}$_{gl}$          & 1.0 ± \textbf{0.03}   & 0.859 ± 0.14 & 0.936 ± 0.65  & 0.963 ± 0.32 & 0.957 ± 0.19 & 0.94 ± 0.34  \\
 \textsc{GraCo}$_{gl}^{reduced}$ & 1.0 ± \textbf{0.03}   & 0.875 ± 0.12 & 0.924 ± 0.64  & 0.949 ± 0.3  & 0.322 ± 0.45 & 0.782 ± 0.39 \\
 \textsc{GraCo}                & 0.978 ± \underline{\textit{0.05}} & 0.876 ± 0.12 & 0.969 ± 0.69  & 0.949 ± 0.3  & 0.961 ± 0.19 & 0.949 ± 0.35 \\
 \textsc{GraCo}$^{reduced}$       & 1.0 ± \textbf{0.03}   & 0.904 ± \underline{\textit{0.1}}  & 0.957 ± 0.67  & 0.939 ± 0.29 & 0.51 ± 0.26  & 0.841 ± 0.35 \\
\hline
\end{tabular}}

\caption{Avg.\ normalized score 
\& mean abs.\ deviation (most indicative, lower is better) from human score for STS
}
\label{tab:av_sts}
\end{table}

\begin{table}[h]
\resizebox{\linewidth}{!}{
\begin{tabular}{@{}lllllll@{}}
\hline
               &   Article &   Aspect &   Co-Hyponymy &   Hyponym &   Omission &   Overall \\
\hline
BLEU           &     0.389 &     \underline{\textit{0.52}} &         0.17  &     0.504 &      0.573 &     0.218 \\
chrF++         &     \underline{\textit{0.611}} &     0.1  &         \underline{\textit{0.68}}  &     0.653 &      0.511 &     0.403 \\
Meteor         &     0.556 &     0.22 &         0.35  &     0.636 &      0.52  &     0.625 \\
\textsc{BERTS}core     &     \textbf{0.722} &     0.1  &         \textbf{0.75}  &     \textbf{0.785} &      \textbf{0.689} &     0.537 \\
\textsc{Smatch}         &     0.333 &     \textbf{1}    &         0.305 &     0.603 &      \underline{\textit{0.591}} &     0.682 \\
\textsc{S$^2$match}        &     0.333 &     \textbf{1}    &         0.305 &     0.603 &      \underline{\textit{0.591}} &     0.682 \\
WLK            &     0.333 &     \textbf{1}    &         0.32  &     0.603 &      0.582 &     \textbf{0.748} \\
WWLK           &     0.333 &     \textbf{1}    &         0.67  &     \underline{\textit{0.769}} &      0.582 &     \underline{\textit{0.712}} \\
\hline
\textsc{GraCo}$_{gl}$         &     0.333 &     0.1  &         0.655 &     0.62  &      0.316 &     0.579 \\
\textsc{GraCo}$_{gl}^{reduced}$ &     0.333 &     0.1  &         0.665 &     0.587 &      0.538 &     0.52  \\
\textsc{GraCo}                &     0.278 &     0.1  &         0.36  &     0.554 &      0.493 &     0.417 \\
\textsc{GraCo}$^{reduced}$ &     0.333 &     0.1  &         0.36  &     0.669 &      \textbf{0.689} &     0.443 \\
\hline
\end{tabular}}
    \caption{Pairwise ranking scores for the STS test cases}
    \label{tab:rank_sts}
\end{table}

\begin{table*}[t]
\centering
\resizebox{0.8\linewidth}{!}{%
\begin{tabular}{l|lll}
                 & \textbf{Best} \& \underline{2nd Best} Ranking Scores & \textbf{Best} \& \underline{2nd Best} MAD & \textbf{Highest} \& \underline{2nd Highest} Correlation w/ Human\\
\midrule
BLEU     &  \underline{Aspect} & & \\
chrF++           & \underline{Co-Hyponymy}, \underline{Article} & \underline{Hyponymy} & \underline{Article}\\
Meteor          &   & \textbf{Co-Hyponymy} & \\
\textsc{BERTS}c      &  \textbf{Hyponymy}, \textbf{Co-Hyponymy}, \textbf{Article}, \textbf{Omission} & \textbf{Omission}, \underline{Hyponymy} & \textbf{Hyponymy}, \textbf{Article}, \textbf{Co-Hyponymy}, \textbf{Omission}\\
\midrule
\textsc{Smatch}         &  \textbf{Aspect}, \underline{Omission} & \textbf{Aspect}, \textbf{Article}, \underline{Omission} & \underline{Omission}\\
\textsc{S$^2$match}      & \textbf{Aspect}, \underline{Omission} & \textbf{Aspect}, \textbf{Article}, \underline{Omission} & \underline{Omission}\\
WLK              & \textbf{Aspect} & \textbf{Aspect}, \textbf{Article}, \underline{Co-Hyponymy} &\\
WWLK        &  \textbf{Aspect}, \underline{Hyponymy} & \textbf{Aspect}, \textbf{Article}, \textbf{Hyponymy} & \underline{Hyponymy}, \underline{Co-Hyponymy}\\
\textsc{GraCo}$_{glo}$ &   & \textbf{Article} &\\
GraCo$_{glo}^{red}$  &  & \textbf{Article} &\\
\textsc{GraCo}  &  & \underline{Article} &\\
GraCo$^{red}$  & \textbf{Omission} & \textbf{Article}, \underline{Aspect} &
\end{tabular}}
    \caption{Overview over \textbf{Best} and \underline{2nd Best} Metric Performances in Ranking Score, MAD and Corr. to Human Scores for the STS dataset.}
    \label{tab:metric_sts}
\end{table*}

\subsection{Experimental Settings}

\subsubsection{Generating sentences from the \textit{Little Prince} AMR corpus.}

We investigated sentences generated from AMRs from the 'Little Prince Corpus'\footnote{\url{https://amr.isi.edu/download.html}} using the AMR-to-text system of \citet{song-etal-2018-graph}. We used their pretrained
\textit{G2S\_silver\_2m} model and validated it on test data from \citet{song-etal-2018-graph}, with a difference of -0.35 points BLEU score. For the 'Little Prince', consisting of  1,562 sentences, we obtained a BLEU score of 13.5.

\begin{table}[hbt!]
\resizebox{\linewidth}{!}{
\begin{tabular}{llcccc}
\toprule
constructional&lexical& SICK & STS&SICK&STS\\
\midrule
Negation && 156 & -\\
Omission && 155 & 15\\
Passive && 78 & -\\
Aspect &&  - & 10\\
Semantic Roles && 8 & -\\
Subordinate Clauses && 69 & -\\
&Antonymy & && 157 & -\\
&Article & && 77 & 6\\
&Hyponymy & && 116 & 11\\
&Co-Hyponymy & && 35 & 20\\
&Partial Synonymy & && 26 & -\\
\midrule
        & & 466 & 25 & 411 & 37\\
Overall & & & & 877 & 62\\
\bottomrule
\end{tabular}}
    \caption{Number of SICK and STS test cases grouped by linguistic phenomena}
    \label{tab:stats}
\end{table}

\subsubsection{Data Statistics} \label{A:statistics}
\begin{figure*}
    \centering
    \includegraphics[scale=0.4]{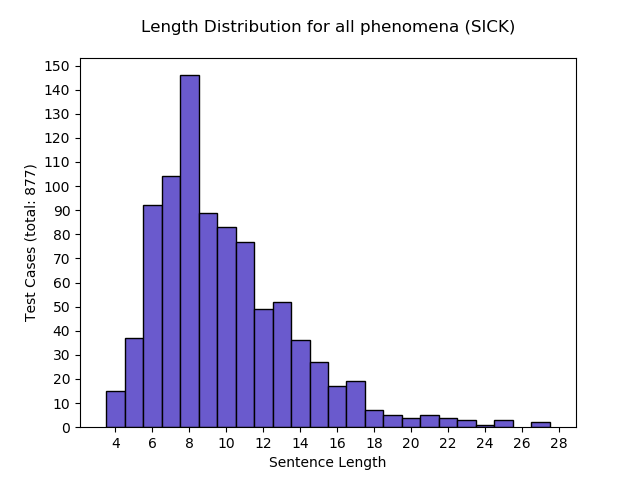}
    \includegraphics[scale=0.4]{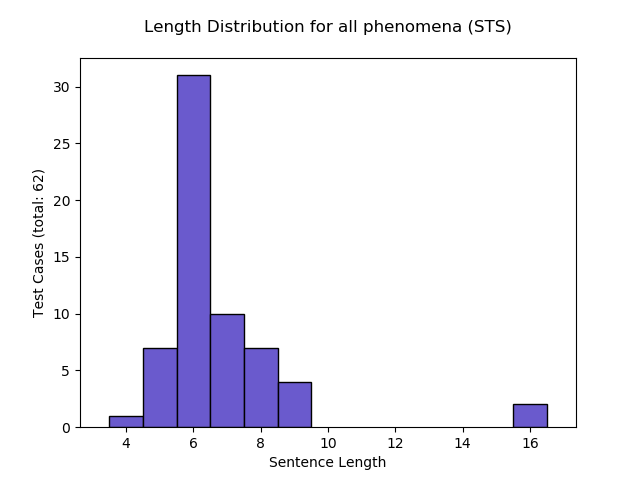}
    \caption{Sentence length distribution for the test cases in the \textit{CheckList} grouped by SICK (left) and STS (right) test cases}
    \label{fig:len_distr}
\end{figure*}

The following figures show the distribution of the human human scores in the \textit{CheckList} for the individual linguistic phenomena. SICK and STS are displayed separately.\\
Fig. \ref{fig:len_distr} further displays the sentence length distribution for SICK and STS.

\begin{figure*}
    \centering
    \includegraphics[scale=0.4]{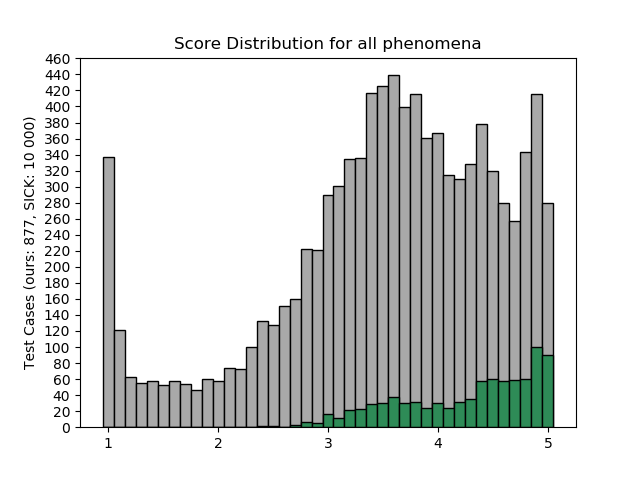}
    \includegraphics[scale=0.4]{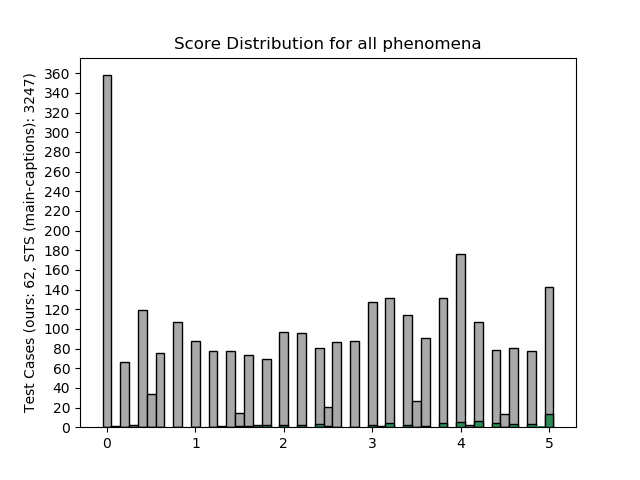}
    \caption{Score distribution for the test cases in the \textit{CheckList} (green) grouped by SICK (left) and STS (right) test cases alongside the distribution of the whole datasets (grey)}
    \label{fig:score_distr}
\end{figure*}

\begin{figure*}
\resizebox{\linewidth}{!}{
\includegraphics{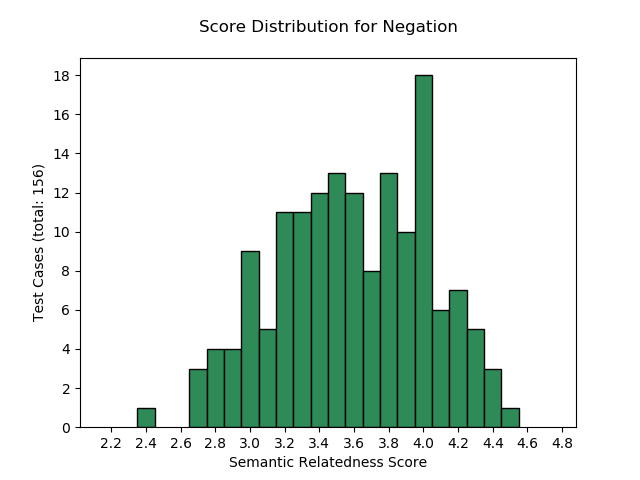}
\includegraphics{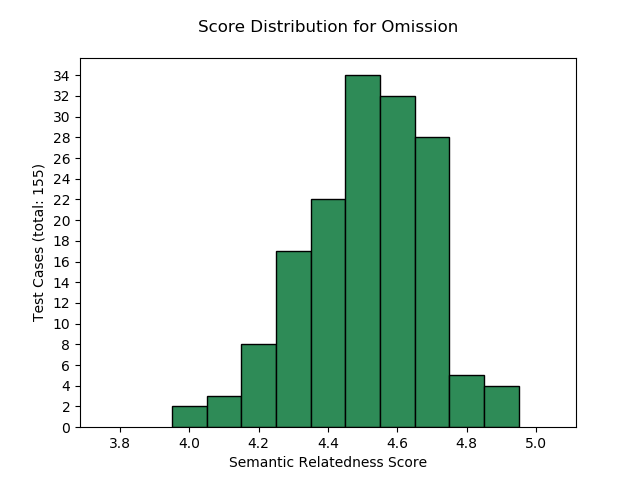}
\includegraphics{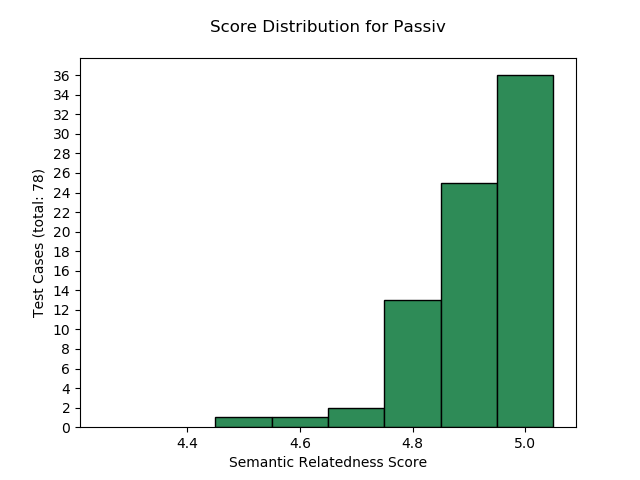}
\includegraphics{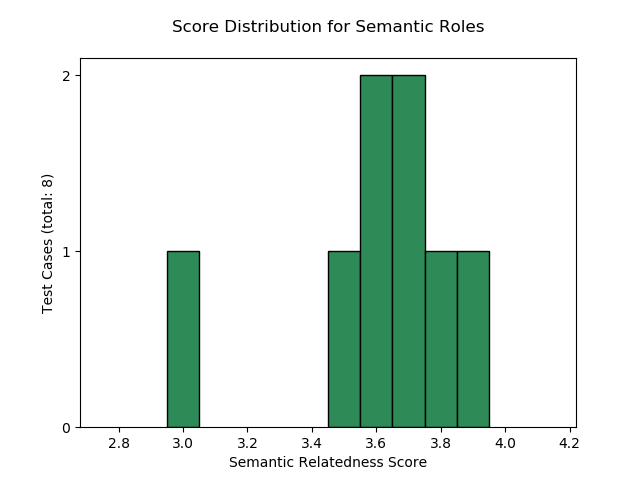}
\includegraphics{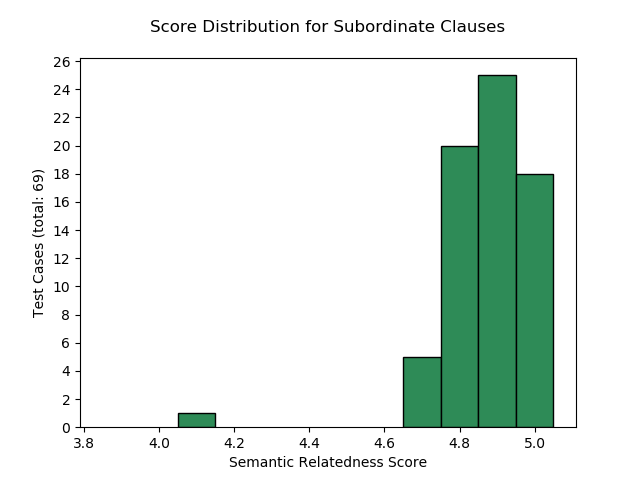}}

\resizebox{\linewidth}{!}{
\includegraphics[scale=0.4]{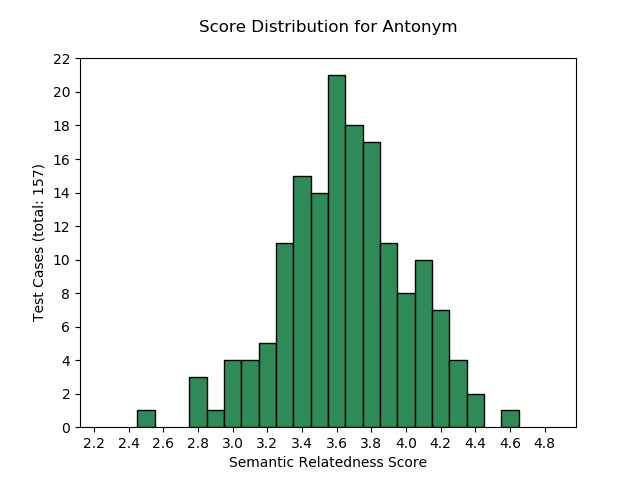}
\includegraphics[scale=0.4]{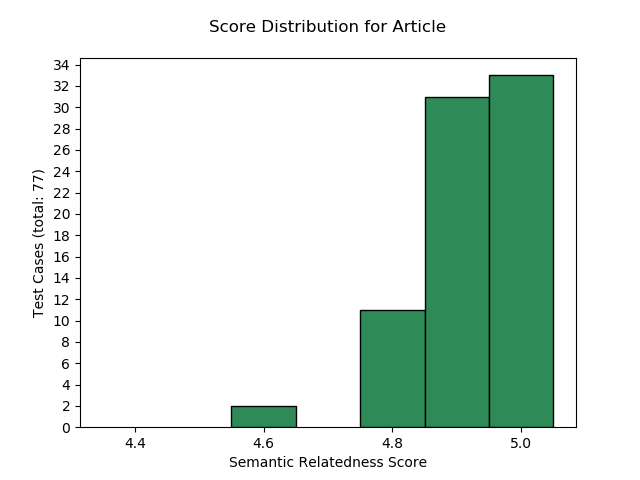}
\includegraphics[scale=0.4]{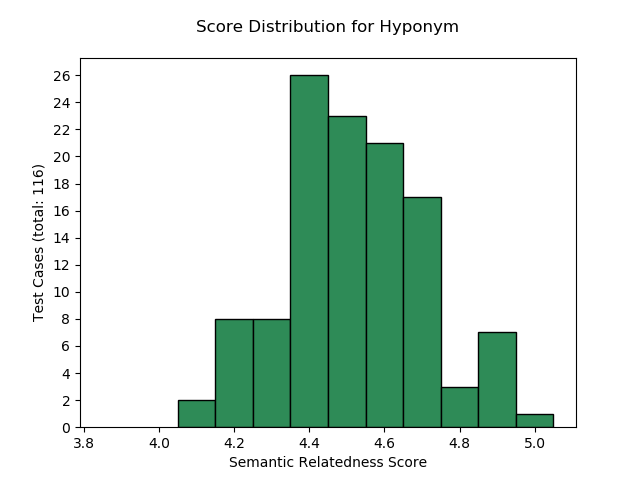}
\includegraphics[scale=0.4]{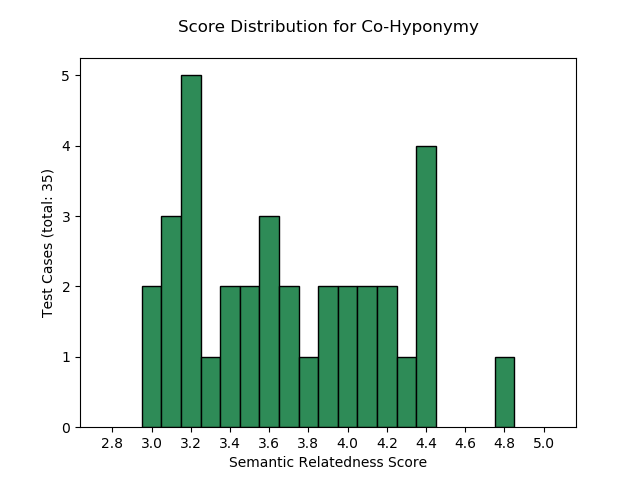}
\includegraphics[scale=0.4]{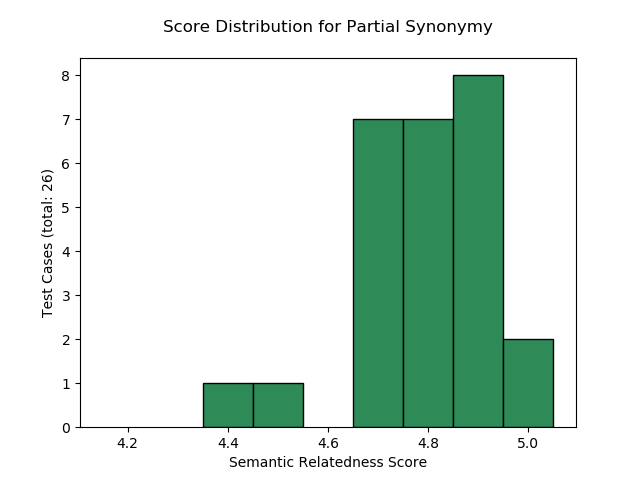}}

\caption{Score distributions for SICK per phenomenon: top: a.) Negation, b. Omission, c. Passive, d. Sem. Roles, e. subord. Clauses; 
bottom: f. Antonymy, g. Article, h. Hymonymy, i. Co-Hyponymy, j. Partial Synonymy.}\label{fig:score_distr_phenomena_sick}

\resizebox{\linewidth}{!}{
\includegraphics{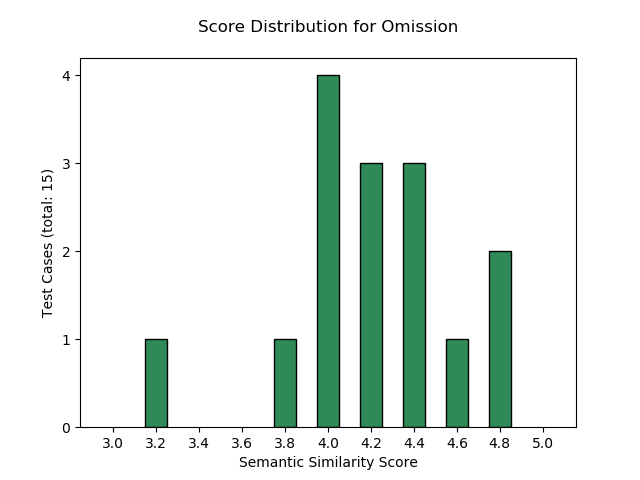}
\includegraphics{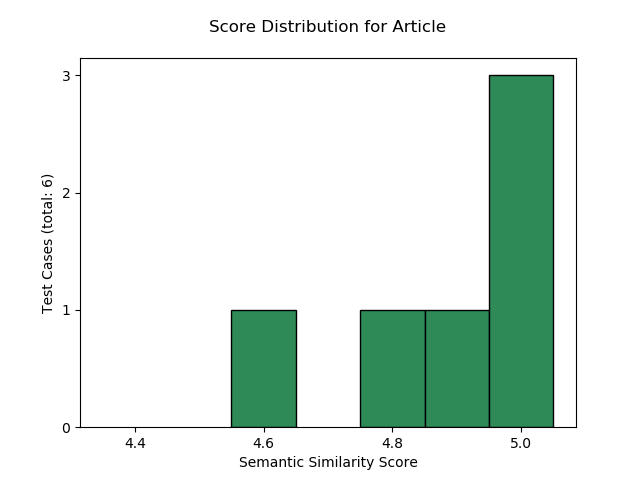}
\includegraphics{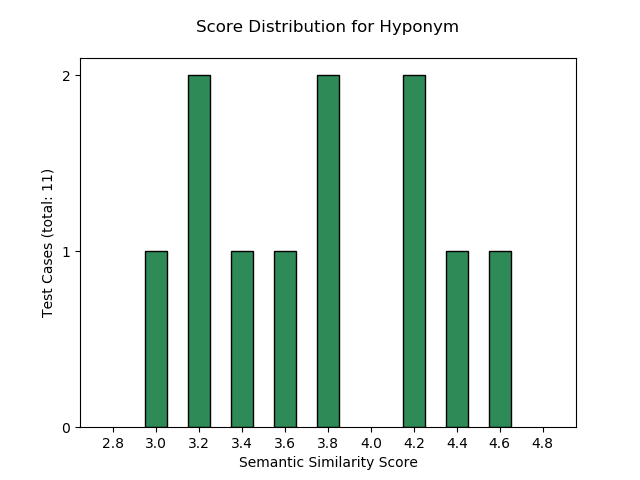}
\includegraphics{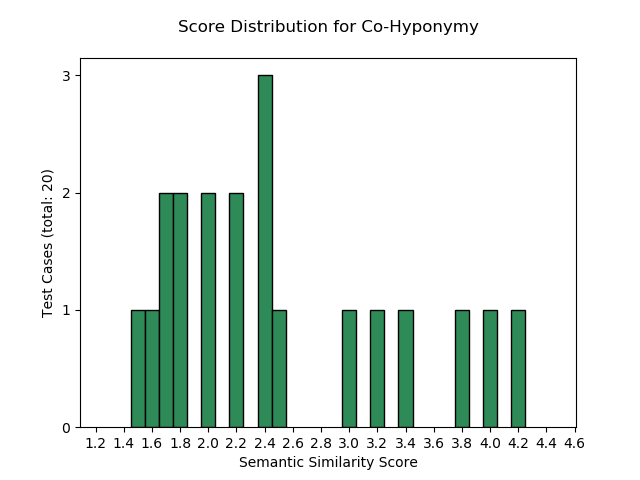}}
\caption{Score distributions for STS per phenomenon: b. Omission, g. Article, h. Hymonymy, i. Co-Hyponymy.}\label{fig:score_distr_phenomena_sts}
\end{figure*}

\subsubsection{Implementation details of metrics} \label{A:hyperparameters}

Here, we list the hyperparameters and libraries employed for the metrics used in the CheckList.

For the text-based metrics, we employ NLTK's implementation for \textbf{BLEU}, where we add the method4 smoothing function \citep{nltk}\footnote{\url{https://www.nltk.org/_modules/nltk/translate/bleu_score.html}}; for \textbf{chrF++} use the sentence-level implementation by \citet{popovic-2015-chrf}, and for 
\textbf{Meteor} the Version 1.5 implementation by \citet{denkowski-lavie-2014-meteor}.

For \citet{BERTscore}'s embedding-based metric \textbf{\textsc{BERTS}core}, we employ the implementation provided by Huggingface\footnote{\url{https://huggingface.co/metrics/bertscore}}.

As for graph-based metrics, we made use of the implementations of \textbf{\textsc{Smatch}} and 
the refined \textbf{\textsc{S$^2$match}} provided by  \citet{opitz-etal-2020-amr}. For \textsc{S$^2$match} we defined a cut-off threshold of 0.9, so that only concepts with a cosine similarity above that threshold would be granted a soft match. Further, the coefficient by which the similarity of differing senses is multiplied was set to 0.95.

For WLK and WWLK we employ the implementation by \citet{opitz-etal-2021-weisfeiler} without any additional hyperparameters.

For the implementation of the \textsc{GraCo}, we used the AMR Alignment tool from JAMR \citep{jamr} to align words from the sentence with concepts in the AMR structure. For concepts that have been successfully aligned, we experimented with contextualized \textsc{BERT} word embeddings, for which we use the \texttt{bert-large-uncased} model with a dimensionality of 1024 \citep{devlin-etal-2019-bert}, and 300 dimensional pretrained GloVe word embeddings \citep{pennington-etal-2014-glove}. In case GloVe may not have seen some inflected word, the embedding of its lemma will be used instead (the lemmata are obtained using the spacy lemmatizer and the \texttt{en\_core\_web\_sm} model). If neither the token nor its lemma is contained in the vocabulary, we generate a zero vector representing an unknown token.

For standardization, given a metric predicts $s=\{s_1,...s_n\}$, where $n$ is the size of the data, we define the standardized score for an example $i$ as $s'_i = \frac{s_i - mean(s)}{std(s)}$. 
Given $s$ as above, the normalized score for an example $i$ is defined as $s'_i = \frac{s_i - min(s)}{max(s) - min(s)}$.

\subsection{Phenomena} \label{A:phenomena}

\subsubsection{Negation}\label{A:negation}
We display two types of negation. In  Fig.\ \ref{fig:neg1} the whole sentence is negated since \texttt{polarity} is attached to the matrix verb. Fig.\ \ref{fig:neg2} shows an AMR where only one constituent in a coordinated sentence is negated.
\begin{figure}[hbt!]
\begin{Verbatim}[commandchars=\\\{\}, frame=single, fontsize=\footnotesize]
        (xv0 / \colorbox{lavender(web)}{exercise-01}  
              :ARG0 (xv1 / man)
              \colorbox{lavender(web)}{:polarity -})
\end{Verbatim}
\caption{AMR for the sentence \textit{The man is not doing excercises.} Semantic relatedness score: 3.8}
\label{fig:neg1}
\end{figure}

\begin{figure}[hbt!]
\begin{Verbatim}[commandchars=\\\{\}, frame=single, fontsize=\footnotesize]
    (xv0 / and
          :op1 (xv1 / walk-01    
                :ARG0 (xv3 / child))
          :op2 (xv2 / \colorbox{lavender(web)}{pull-up-07}        
                :ARG1 (xv5 / jeep-01)    
                \colorbox{lavender(web)}{:polarity -})
\end{Verbatim}
    \caption{AMR for the sentence \textit{A child is walking and a jeep is not pulling up.} Semantic relatedness score: 3.5}
    \label{fig:neg2}
\end{figure}

\subsubsection{Omission and Hallucination} \label{A:hallucination}
Fig.\ \ref{fig:om1} demonstrates the AMR of the sentence \textit{The man is \underline{cautiously} operating a stenograph}. The adverb is realized by the use of the role \texttt{:manner}. The sentence \textit{The man is operating a stenograph} would look the same, except that the red-colored branch would not exist.
Since concepts can be described in various ways, some words may be represented by more than one branch which would lead to more than two triples that don't have a counterpart.
The omission of a prepositional phrase often resembles the omission of adjectives or adverbs, especially for phrases that can be realized by so-called \enquote{none-core-roles} such as \texttt{destination}, \texttt{location} or \texttt{medium}, hence, within one branch. As described in section A.3, prepositions, however, can be realized in various ways. The omission of a prepositional expression might therefore concern only one branch, but can also concern multiple branches like in Fig. \ref{fig:om2}.

\begin{figure*}[t!]
\begin{Verbatim}[commandchars=\\\{\}, fontsize=\footnotesize, frame=single]
(xv0 / play-11                           (xv0 / play-11  
     :ARG0 (xv2 / man)                        :ARG0 (xv2 / man)   
     :ARG1 (xv1 / flute                       :ARG1 (xv1 / flute 
          \colorbox{palechestnut}{:consist-of (xv3 / bamboo)}))             \colorbox{palechestnut}{:ARG1-of (xv3 / make-01}     
                                                          \colorbox{palechestnut}{:ARG2 (xv4 / bamboo))})) 
\end{Verbatim}
    \caption{AMR structures for the sentence pair \textit{A man is playing a bamboo flute -- A man is playing a flute made of bamboo} Semantic relatedness score: 4.9}
    \label{fig:sub2}
\end{figure*}
\begin{figure}[hbt!]
\begin{Verbatim}[commandchars=\\\{\}, frame=single, fontsize=\footnotesize]
  (xv0 / operate-01
      :ARG0 (xv2 / man)
      :ARG1 (xv1 / stenograph)
          \colorbox{palechestnut}{:manner (xv3 / cautious-02)})
\end{Verbatim}
    \caption{Gold AMR for the sentence \textit{A man is \\cautiously operating a stenograph.} Semantic\\relatedness score: 4.5}
    \label{fig:om1}
\end{figure}
\begin{figure}[hbt!]
\begin{Verbatim}[commandchars=\\\{\}, frame=single, fontsize=\footnotesize]
    (xv0 / attack-01
        :ARG0 (xv2 / dog
            :mod (xv3 / brown))
        :ARG1 (xv1 / animal)
        \colorbox{palechestnut}{:location (xv4 / in-front-of}
            \colorbox{palechestnut}{:op1 (xv5 / man))})
\end{Verbatim}
    \caption{Gold AMR for the sentence \textit{The brown\\ dog is attacking an animal in front of the man.}}
    \label{fig:om2}
\end{figure}

\subsubsection{Passive}
Since AMR aims to capture the events of a sentence and not necessarily its \textit{point of view}, AMR structures of an active-passive sentence pair do not differ at all.

\subsubsection{Semantic and Syntactic Role Switch} \label{A:roleswitch}
The AMRs in Fig.\ \ref{fig:ssrs} show that semantic and syntactic role switch is expressed by switching the \texttt{:ARG} roles. This results in the pair of AMRs differing in two triples.
\begin{figure}[h]
    \centering
\begin{Verbatim}[commandchars=\\\{\}, frame=single, fontsize=\tiny]
(xv0 / follow-02                (xv0 / follow-02
    \colorbox{palechestnut}{:ARG0 (xv1 / turtle)}           \colorbox{palechestnut}{:ARG0 (xv2 / fish)}
    \colorbox{palechestnut}{:ARG1 (xv2 / fish)})            \colorbox{palechestnut}{:ARG1 (xv1 / turtle)})
\end{Verbatim}
    \caption{AMR structures of the sentence pair \textit{The turtle is following the fish. -- The fish is following the turtle.} Semantic relatedness score: 3.8}
    \label{fig:ssrs}
\end{figure}
\subsubsection{Subordinate Clauses} \label{A:sub1}
In \S \ref{sub_phen} we already discussed \textit{inverse roles} for relative clauses when the relativizer is dependet on a verb. For attributive adjectives on the other hand, AMR structures should look the same. This is demonstrated by the AMR representations for \textit{A black bird is sitting on a dead tree} and \textit{A bird, which is black, is sitting on a dead tree} in Fig. \ref{fig:sub1}. Fig. \ref{fig:sub2} displays a sentence pair featuring a noun compound expansion.

\begin{figure}[h]
    \centering
\begin{Verbatim}[commandchars=\\\{\}, frame=single, fontsize=\footnotesize]
    (xv0 / sit-01
        :ARG1 (xv1 / bird
            \colorbox{lavender(web)}{:ARG1-of (xv3 / black-04)})
        :ARG2 (xv2 / tree
            :ARG1-of (xv4 / die-01)))
\end{Verbatim}
    \caption{AMR structure for the sentence pair \textit{A black bird is sitting on a dead tree. -- A bird, which is black, is sitting on a dead tree.} Semantic relatedness score: 5.0}
    \label{fig:sub1}
\end{figure}

\subsubsection{Article} \label{A:article}
\citet{banarescu-etal-2013-abstract} specifically state that \enquote{AMR does not represent inflectional morphology for tense and number, and [...] omits articles}. 

\subsubsection{Antonymy} \label{A:antonymy}
In Fig. \ref{fig:ant1}, we see two AMR graphs for a sentence pair exhibiting an antonymous relation between \textit{young} and \textit{old}. The antonymy is realized by mapping the differing concepts to the variable \texttt{xv3} respectively.

\begin{figure}[h]
    \centering
\begin{Verbatim}[commandchars=\\\{\}, fontsize=\tiny, frame=single]

(xv0 / talk-01                  (xv0 / talk-01
    :ARG0 (xv1 / man                :ARG0 (xv1 / man
        :mod \colorbox{palechestnut}{(xv3 / young)})            :mod \colorbox{palechestnut}{(xv3 / old)})
    :ARG2 (xv2 / leaf))             :ARG2 (xv2 / leaf))
            
\end{Verbatim}
    \caption{AMR structures for the sentence pair \textit{A \colorbox{lavender(web)}{young} man is talking to a leaf. -- An \colorbox{lavender(web)}{old} man is talking to the leaf.} Semantic relatedness score: 3.915}
    \label{fig:ant1}
\end{figure}

Another way of realizing antonymy between adjectives in an AMR graph is adding the feature \texttt{:polarity -} to the branch of the adjective's concepts which inverts its meaning.

\subsubsection{Hyperonymy, Hyponymy and Co-Hyponymy}
An AMR structure of two sentences displaying a sub- or superset relation would differ merely in the concepts mapped to the corresponding variable as demonstrated in Fig. \ref{fig:hyp}. This is also true for co-hyponymy.

\begin{figure}[h]
    \centering
\begin{Verbatim}[commandchars=\\\{\}, frame=single, fontsize=\tiny]
(xv0 / run-02                   (xv0 / run-02    
    :ARG0 \colorbox{palechestnut}{(xv2 / squirrel)}        :ARG0 \colorbox{palechestnut}{(xv2 / animal)}
    :ARG1 (xv1 / circle))           :ARG1 (xv1 / circle))
\end{Verbatim}
    \caption{AMR structures for the sentence pair \textit{A squirrel is running in circles. -- An animal is running in circles.} Semantic relatedness score: 4.4}
    \label{fig:hyp}
\end{figure}

\end{document}